\documentclass{fairmeta} % 使用单栏排版为主，twocolumn 可选
\usepackage[utf8]{inputenc}
\usepackage{url}
\usepackage{verbatim}
\usepackage{xspace}
\usepackage{enumitem}
\usepackage{textcomp}
\usepackage{comment}
\usepackage{titlesec}
\usepackage{tocloft}
\usepackage{float}

\usepackage{graphicx}
\usepackage{wrapfig}
\usepackage{adjustbox}
\usepackage{booktabs}
\usepackage{multirow}
\usepackage{array}
\usepackage{colortbl}
\usepackage{tabularx}
\usepackage{subcaption}
\usepackage{algpseudocode}

\newcolumntype{x}[1]{>{\centering\arraybackslash}p{#1pt}}
\newcolumntype{y}[1]{>{\raggedright\arraybackslash}p{#1pt}}
\newcolumntype{z}[1]{>{\raggedleft\arraybackslash}p{#1pt}}

\newlength\savewidth

\usepackage{amsmath, amssymb}
\usepackage{bm} % 可选，用于粗体向量
\usepackage{pifont}
\usepackage{bbding}

\usepackage{xcolor}

\definecolor{adptorange}{RGB}{248, 205, 172}
\definecolor{cmpblue}{RGB}{189, 215, 238}
\definecolor{our_red}{RGB}{232,157,160}
\definecolor{our_blue}{RGB}{136,206,230}
\definecolor{our_orange}{RGB}{246,200,168}
\definecolor{our_green}{RGB}{178,211,164}
\definecolor{attn_code0}{RGB}{247,215,200}
\definecolor{attn_code1}{RGB}{238,169,139}
\definecolor{mlp_code0}{RGB}{204,201,221}
\definecolor{mlp_code1}{RGB}{102,95,153}
\definecolor{token_blue}{RGB}{84, 120, 140}
\definecolor{codeblue}{rgb}{0.25, 0.5, 0.5}
\definecolor{codekw}{rgb}{0.35, 0.35, 0.75}
\definecolor{green}{HTML}{009000}
\definecolor{red}{HTML}{ea4335}
\definecolor{lightgray}{gray}{0.95}
\definecolor{darkblue}{rgb}{0.1,0.1,0.6}
\definecolor{darkgreen}{rgb}{0.0,0.5,0.0}
\definecolor{darkred}{rgb}{0.6,0.1,0.1}
\definecolor{paleblue}{RGB}{250, 253, 255}
\definecolor{palegreen}{RGB}{250, 255, 250}
\definecolor{palecream}{RGB}{255, 253, 250}

\definecolor{mycolback1}{RGB}{245,246,250}      % box 背景：很浅的灰蓝
\definecolor{mycolframe1}{RGB}{180,188,205}     % box 边框：中等灰蓝
\definecolor{mycoltitleback1}{RGB}{120,160,200} % 标题栏背景：偏深蓝
\definecolor{mycoltitleframe1}{RGB}{120,160,200}% 标题栏边框

\usepackage[most]{tcolorbox}

\tcbset{
  myboxstyle/.style={
    enhanced,
    colback=mycolback1,      % 背景：浅灰蓝，和 lightgray 比较协调
    colframe=mycolframe1,    % 边框：中度灰蓝
    coltitle=black,
    sharp corners,
    boxrule=0.8pt,
    left=8pt,
    right=8pt,
    top=8pt,
    bottom=8pt,
    arc=2pt,
    attach boxed title to top left={yshift=-2mm, xshift=2mm},
    boxed title style={
        colback=mycoltitleback1,   % 标题栏背景：稍深的蓝色
        colframe=mycoltitleframe1, % 标题栏边框
        arc=2pt,
        boxrule=0pt,
        left=6pt,
        right=6pt,
        top=2pt,
        bottom=2pt,
        fontupper=\bfseries\color{white}, % 标题栏文字：白色
    },
    drop fuzzy shadow={gray!30!white}, % 柔和阴影
  }
}

\definecolor{mycolback}{RGB}{240,245,255}      % 浅蓝背景
\definecolor{mycolframe}{RGB}{140,170,210}     % 蓝色边框
\definecolor{mycoltitleback}{RGB}{100,140,190} % 标题栏背景蓝
\definecolor{mycoltitleframe}{RGB}{100,140,190}

\tcbset{
  drfstyle/.style={
    enhanced,
    colback=mycolback,
    colframe=mycolframe,
    sharp corners,
    boxrule=0.8pt,
    arc=2pt,
    left=8pt, right=8pt, top=8pt, bottom=8pt,
    drop fuzzy shadow={gray!30!white},
    attach boxed title to top left={yshift=-2mm, xshift=2mm},
    boxed title style={
      colback=mycoltitleback,
      colframe=mycoltitleframe,
      fontupper=\bfseries\color{white},
      arc=2pt,
      top=2pt, bottom=2pt, left=6pt, right=6pt
    }
  }
}

\usepackage{algorithm}
\usepackage{listings}
\usepackage{caption}
\usepackage{float} % for [H]

\lstdefinestyle{Pytorch}{
    language = Python,
    backgroundcolor = \color{white},
    basicstyle = \fontsize{9pt}{8pt}\selectfont\ttfamily\bfseries,
    columns = fullflexible,
    aboveskip=1pt,
    belowskip=1pt,
    breaklines = true,
    captionpos = b,
    commentstyle = \color{codeblue},
    keywordstyle = \color{codekw},
}

\usepackage{hyperref}

\renewcommand{\paragraph}[1]{\vspace{1.25mm}\noindent\textbf{#1}}

\makeatletter
\DeclareRobustCommand\onedot{\futurelet\@let@token\@onedot}
\def\@onedot{\ifx\@let@token.\else.\null\fi\xspace}

\makeatother

\usepackage{tcolorbox}
\newcommand{\ours}{\textbf{{Lumos-Nexus}}\xspace}
\newcommand{\ourbridge}{{Lumos-Nexus}\xspace}

\tcbuselibrary{listings, skins, breakable}

\tcbset{
  myboxstyle/.style={
    boxrule=1pt,
    boxsep=2pt,
    colback=gray!5,
    coltext=black, % 新增：设置文字为黑色
    fontupper=\scriptsize,  % 控制 box 内普通文字的字体
    fonttitle=\color{black},
    title=Prompt design for entity words retrieval and subject-attribute matching in Qwen2.5-VL,
    coltitle=black,
    colframe=black,
    colbacktitle=blue!10,
    breakable
  }
}

\tcbset{
  myboxstyle2/.style={
    boxrule=1pt,
    boxsep=2pt,
    colback=gray!5,
    coltext=black, % 新增：设置文字为黑色
    fontupper=\scriptsize,  % 控制 box 内普通文字的字体
    fonttitle=\color{black},
    title= Instruction design for inpainting model assessment in GPT-4o,
    coltitle=black,
    colframe=black,
    colbacktitle=blue!10,
    breakable
  }
}

\newtcblisting{promptbox}[1][]{
  listing only,
  myboxstyle, % 应用你定义的 box 样式
  listing options={
    language=Python,
    basicstyle=\ttfamily\scriptsize, 
    rulecolor=\color{gray},            
    breaklines=true,
  },
  #1
}
\newtcblisting{promptbox2}[1][]{
  listing only,
  myboxstyle2, % 应用你定义的 box 样式
  listing options={
    language=Python,
    basicstyle=\ttfamily\scriptsize, 
    rulecolor=\color{gray},            
    breaklines=true,
  },
  #1
}
\usepackage{caption}
\title{Lumos-Nexus: Efficient Frequency Bridging with Homogeneous Latent Space for Video Unified Models}

\author{%
  Jiazheng Xing$^{*1,4,2}$, 
  Hangjie Yuan$^{*\ddagger2,3,1}$, 
  Lingling Cai$^{1}$, 
  Xinyu Liu$^{5}$,
  Yujie Wei$^{6}$, 
  Fei Du$^{2,3}$, 
  Tao Feng$^{7}$, 
  Hai Ci$^{4}$, 
  Jiasheng Tang$^{2,3}$,
  Weihua Chen$^{\dagger2,3}$,
  Fan Wang$^{2}$,
  Yong Liu $^{\dagger1}$\\
  \vspace{.1cm}
  \small{
  $^1$Zhejiang University,  
  $^2$DAMO Academy, Alibaba Group,
  $^3$Hupan Lab,
  $^4$National University of Singapore,\\
  $^5$Hong Kong University of Science and Technology,
  $^6$Fudan University,
  $^7$Tsinghua University \\}
  \small{* Equal contribution, $\ddagger$ Project lead, $^\dagger$ Corresponding authors.}\\
  \vspace{.1cm}
  \texttt{jiazhengxing@zju.edu.cn, kugang.cwh@alibaba-inc.com, yongliu@iipc.zju.edu.cn}  \\

Project Page: \url{https://jiazheng-xing.github.io/nexus-lumos-home/}

}

\abstract{
Connector-based video unified models have demonstrated strong capability in instruction-grounded video synthesis, but integrating a large high-fidelity generator into the unified training loop is computationally prohibitive, limiting achievable visual quality.
We therefore propose \ours, a training-efficient unified video generation framework that facilitates the development of strong reasoning-driven generation capabilities while significantly enhancing visual fidelity. \ourbridge adopts a two-stage design: 1) During training, only a lightweight generator is aligned with the understanding block to learn to take in reasoning-driven semantic control. 2) During inference, we introduce Unified Progressive Frequency Bridging (UPFB) to progressively hand off generation to a high-capacity pretrained generator in the shared latent space, enabling coarse-to-fine refinement and producing high-fidelity videos without compromising reasoning quality. {To fill the gap in reasoning-driven video generation benchmarks, we introduce VR-Bench, which assesses a model’s capability to translate inferred intent into coherent and semantically aligned video content.}
Extensive experiments demonstrate that \ourbridge achieves substantial gains in visual realism and temporal coherence on VBench, while exhibiting strong reasoning-based generative performance on VR-Bench.
}
\date{\today}

\begin{document}
\maketitle

    \section{Introduction}
\label{sec:intro}

Unified models~\cite{deng2025emerging, pan2025transfer, chen2025blip3, tong2025metamorph, xie2024show, ai2025ming, ge2024seed, wu2024next, dong2023dreamllm} have emerged as a promising paradigm that integrates multimodal understanding and generative modeling into a unified system, offering strong potential for the two processes to mutually reinforce one another. In particular, it has been shown that the understanding block supplies structured semantic priors to the visual generator, enabling it to interpret complex instructions and produce outputs aligned with coherent logical intent.
Extending this paradigm to video modeling is particularly crucial, as video is not a static visual form but a sequence of events unfolding over time. Generating videos requires maintaining temporal consistency, causal progression, and coherent motion, which inherently demands stronger reasoning than image generation. As a result, video unified models~\cite{xie2025show,tan2025omni,wei2025univideo,luo2025univid} must effectively bridge high-level semantic understanding with temporally consistent generation, making the interaction between understanding and generation even more crucial.

\begin{figure*}[t!]
  \centering
  \includegraphics[width=0.95\linewidth]{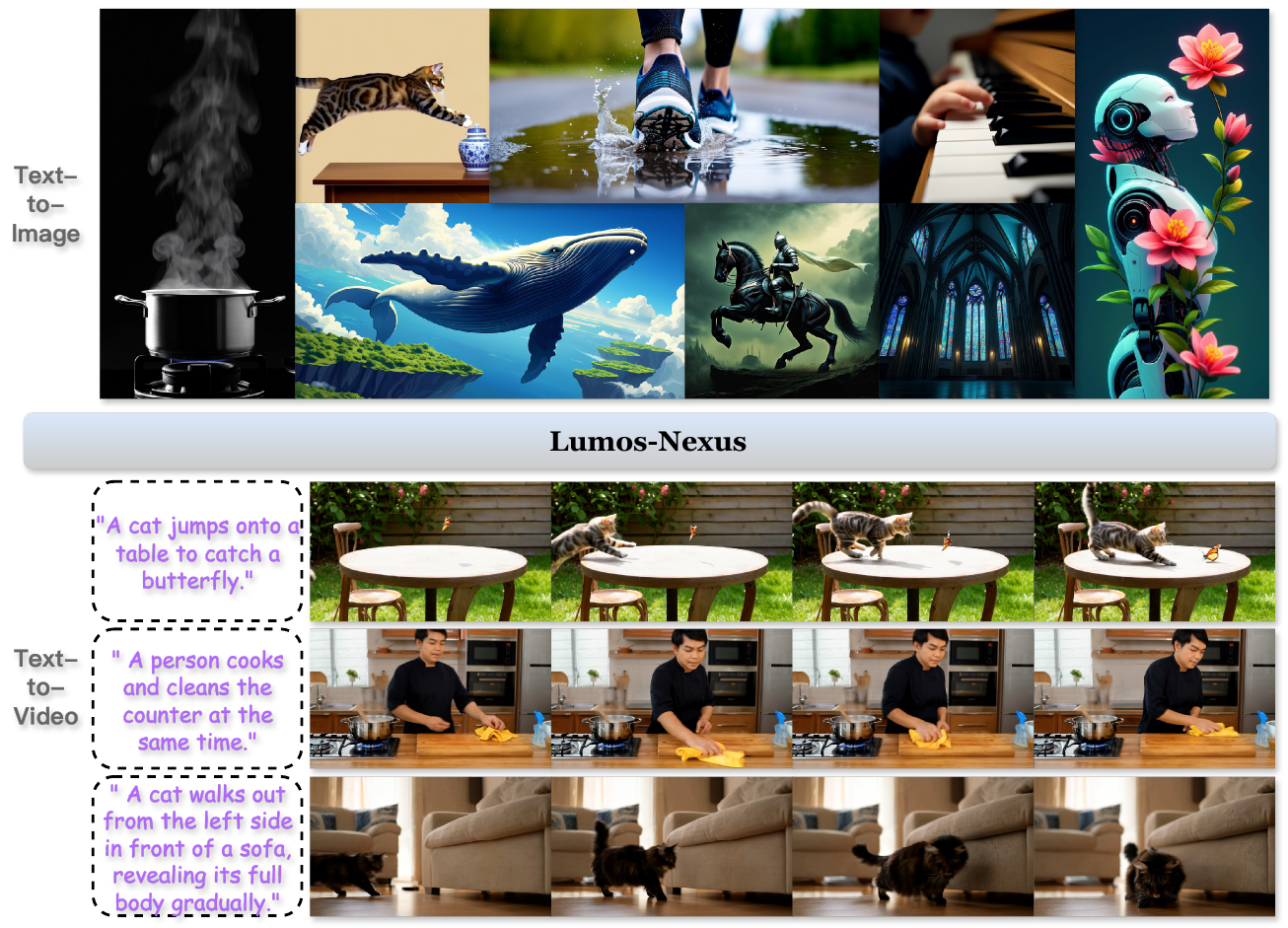}
  \vspace{-5pt}
  \caption{Visualization of examples generated by \ours. Our \ourbridge supports both text-to-image (T2I) and text-to-video (T2V) generation, demonstrating reasoning-aware generation with high-fidelity visual outputs.}
  \label{fig:teaser}
  \vspace{-15pt}
\end{figure*}

% Video unified models can be broadly divided into parallel-based~\cite{xie2025show, luo2025univid, wei2025univideo} and connector-based~\cite{tan2025omni} architectures. 
% \hangjie{
% From the architectural perspective, video unified models can be broadly divided into parallel-based~\cite{xie2025show, luo2025univid, wei2025univideo} and connector-based~\cite{tan2025omni} models. 
% }
{From the architectural perspective, video unified models can be broadly divided into the joint-attention~\cite{xie2025show, luo2025univid, wei2025univideo} and connector-based~\cite{tan2025omni} models. }
Joint-attention models enable long-context interaction between multimodal understanding and generation through shared self-attention, offering stronger scalability but requiring substantially more extensive training. In contrast, connector-based models introduce an explicit connector, which can be designed to align the understanding block's representation into the visual generator's condition injection space, {thus avoiding joint optimization of both the understanding block and the generation block. }
% thus reducing the need to retrain the generator extensively. 
% \hangjie{thus avoiding joint optimization of both the understanding block and the generation block}
% However, despite their efficiency advantages, connector-based video unified models still face prohibitive computational cost when high-quality video generation requires scaling up to large diffusion generators—making it challenging to simultaneously achieve strong semantic alignment and high visual fidelity in practice. 
{However, despite this decoupled design, connector-based video unified models still require prohibitive fine-tuning overhead used for aligning understanding output with the generation input, due to the substantially large-scale diffusion generator (\textit{e.g.}, Wan2.1-14B~\cite{wang2025wan})—making it challenging to simultaneously achieve strong semantic alignment and high visual fidelity in practice.}
% \hangjie{
% However, despite this decoupled design, connector-based video unified models still require prohibitive fine-tuning overhead used for aligning understanding output with the generation input, due to the substantially large-scale diffusion generator (\textit{e.g.}, Wan2.1 14B [cite])
% % , to ensure quality generation results
% }

Given practical computational constraints, we focus on the connector-based paradigm. However, directly fine-tuning a large diffusion generator within this framework remains computationally expensive. 
To address this, we explore whether a smaller diffusion generator—\textit{homogeneous in latent space with the larger generator}—can be used during training instead. 
Since fine-tuning in unified models does not alter the latent representation space of the diffusion backbone, maintaining a shared latent space between the two generators establishes a foundation for bridging them during inference. 
Our key idea is to allow the small generator to learn how to absorb and encode high-level semantic priors from the understanding block within the unified training loop. 
Then, at inference time, this small generator acts as a semantic initiator, transforming the understanding-derived semantic representation into coherent structural priors that can be seamlessly inherited by the large generator. The large generator, pretrained on extensive video data, subsequently contributes its stronger high-fidelity synthesis ability and further reinforces the execution of reasoning-driven semantics. In this way, we achieve video generation that is both semantically accurate and visually high-quality, without requiring full-scale training of the large generator inside the unified model.

To build a training-efficient and high-quality unified video generation framework, we propose \ours, which significantly enhances visual fidelity while strengthening reasoning-driven generative capability, as shown in Fig~\ref{fig:teaser}. \ourbridge is structured in two stages, disentangling the acquisition of semantic alignment from the process of high-fidelity video synthesis. (1) \textit{Training stage}. We align only a lightweight diffusion generator with the understanding block so that it learns to transform semantic and reasoning cues into structured generative signals. (2) \textit{Inference stage}. We introduce Unified Progressive Frequency Bridging (UPFB), which progressively transfers the generative responsibility from the lightweight generator to a pretrained high-capacity generator operating in the shared homogeneous latent space. This controlled handoff produces a natural coarse-to-fine refinement process, enabling high-fidelity, temporally coherent videos while further reinforcing reasoning behavior learned during training. Meanwhile, to address the lack of benchmarks for reasoning-driven video generation and to verify that our method preserves the unified model’s reasoning ability, we introduce \textbf{VR-Bench}, {which systematically evaluates the alignment between inferred intent and generated video content across eight dimensions spanning physical-world reasoning, commonsense reasoning and embodied interactions.}
% which evaluates the alignment between inferred intent and generated video content across multiple dimensions.
% \hangjie{You should be more specific on ``multiple dimensions" since it is also one of your contributions.}
Extensive experiments demonstrate that \ourbridge achieves substantial improvements in visual realism and temporal coherence on VBench, while maintaining strong reasoning-based generative performance on VR-Bench. The contribution of our \ourbridge can be summarized as follows:

\begin{itemize}
\item We propose \ourbridge, a training-efficient unified video generation framework that aligns reasoning-guided semantics using a lightweight generator during training, and employs  Unified Progressive Frequency Bridging (UPFB) at inference to progressively hand off generation to a high-capacity generator, achieving high-fidelity video synthesis while further enhancing reasoning capability.

\item We introduce VR-Bench, which evaluates the alignment between inferred intent and generated video content across multiple dimensions in video generation models.

\item Extensive experiments show that \ourbridge significantly improves visual realism and temporal coherence on VBench, while maintaining strong reasoning-guided generation on VR-Bench.

\end{itemize}

\section{Related Works}
\label{sec:related_works}
\noindent \textbf{Video Generation Models.}
Recent progress in video generation has been driven primarily by advances in both autoregressive~\cite{yan2021videogpt, hong2022cogvideo, kondratyuk2023videopoet, liu2024mardini, teng2025magi, yuan2025lumos, jin2024pyramidal, li2024arlon, deng2024autoregressive, wang2024loong} and diffusion-based~\cite{wu2023tune,singer2022make, blattmann2023stable, wang2023modelscope, lu2023vdt, yang2024cogvideox, ma2024latte, khachatryan2023text2video, kong2024hunyuanvideo, chen2024videocrafter2, zheng2024open, wang2025wan, team2025longcat, zhang2025show, qiu2025skyreels, xing2026lumosx} modeling paradigms. Autoregressive approaches transform videos into discrete token sequences and generate them step-by-step through large transformers, enabling explicit temporal modeling but typically suffering from high inference latency and accumulated error over long sequences. Diffusion-based video models instead synthesize video by denoising latent representations, and have demonstrated strong temporal smoothness and visual quality. Early diffusion-based methods~\cite{wu2023tune, wang2023modelscope, chen2024videocrafter2, khachatryan2023text2video} largely extended 2D U-Net architectures used for text-to-image generation to the video domain, which limited temporal expressiveness. More recent architectures~\cite{lu2023vdt, yang2024cogvideox, ma2024latte, kong2024hunyuanvideo, zheng2024open, wang2025wan, team2025longcat, zhang2025show} employ diffusion transformers (DiTs) with spatiotemporal attention, improving motion consistency and scalability to high resolutions.

\noindent \textbf{Video Unified Models.}
Unified models~\cite{wang2024emu3, deng2025emerging, pan2025transfer, chen2025blip3, zhou2024transfusion, tong2025metamorph, xie2024show, ai2025ming, yang2025mmada, wu2025janus, chen2025janus} integrate multimodal understanding and visual generation within a single framework, where the two components mutually reinforce one another, and existing architectures are commonly categorized into autoregressive-based~\cite{wang2024emu3, wu2025janus}, diffusion-based~\cite{yang2025mmada}, and hybrid formulations~\cite{zhou2024transfusion, deng2025emerging, pan2025transfer, chen2025blip3, tong2025metamorph, xie2024show, ai2025ming}. Video unified models~\cite{xie2025show, luo2025univid, wei2025univideo, chen2025univid, tan2025omni} extend the unified modeling paradigm to the video domain, enabling semantic grounding and logical reasoning within video synthesis. In terms of fusion strategy between understanding and generation, video unified models can be grouped into joint-attention~\cite{xie2025show, luo2025univid, wei2025univideo, chen2025univid} designs, which perform long-context interaction via shared self-attention, and connector-based~\cite{tan2025omni} designs, which inject understanding features into the generator through a dedicated connector. The former offers stronger scalability but demands heavy training cost, while the latter is more computation-efficient yet still faces difficulty when scaling to large diffusion generators for high-fidelity video synthesis. Considering practical computational constraints, we focus on the connector-based paradigm in this work.
\section{Methods}
\label{method}

\subsection{Preliminary}
% 我们关注unified model中理解模型促进生成的能力。过去的工作已经证明了随着多模态理解模型的引入，从中提取的世界知识注入到生成模型中，能有助于生成模型去实现智能生成，即面对复杂或者没那么直接的instruction，模型能够理解instruction的语义进行逻辑推理，基于推理结果完成生成。在这些image/video unified model中无论才用何种结构都需要面临将理解模型的知识注入进生成模型中，如下：
% (给一个公式描述)
% 在这一注入过程中为了使生成器有效接受理解模型输出的世界模型，任何结构都无法避免训练生成器。尤其是在视频Unified Model中，训练生成器的成本较为高昂。While directly incorporating a large generator into the entire end-to-end understanding-to-generation loop of the Video Unified Model training framework is computationally prohibitive, the ability of a larger generator to produce high-fidelity visual details is nevertheless highly appealing.
% We focus on the capability of unified models to leverage pretrained vision–language models (VLMs, i.e., used for understanding) in support of conditional generation.
In connector-based unified models, prior studies~\cite{pan2025transfer, tong2025metamorph, chen2025blip3, tan2025omni} have shown that extracting world knowledge from pretrained vision–language models (VLMs) used for understanding and injecting it into diffusion transformers can effectively enhance conditional generation. Especially in text-to-video (T2V) generation, when confronted with complex or indirectly specified textual instructions, such unified models can interpret semantic intent, perform logical reasoning, and subsequently synthesize visual content grounded in the inferred reasoning process.
For connector-based image or video unified models, a crucial step is injecting the output of the understanding block as a conditioning signal into the generation block, which can be formally expressed as:
\begin{equation}
\mathbf{c}_{\mathcal{U}} = f_{\mathcal{C}}\left({\mathcal{U}}\left(x; \theta_{\mathcal{U}}\right); \theta_{\mathcal{C}}\right), 
\qquad
\hat{y}_\mathcal{G} \sim p_{\theta_{\mathcal{G}}}(y \mid \mathbf{c}_{\mathcal{U}}),
\label{eq:conditioning}
\end{equation}
where $x$ denotes the textual instruction input. 
${\mathcal{U}}$ serves as the {understanding block} parameterized by $\theta_{\mathcal{U}}$ that processes the textual input into a world-knowledge-grounded semantic representation, which is further transformed by the {connector} $f_{\mathcal{C}}(\ \cdot \ ;\theta_{\mathcal{C}})$ into a generator-compatible conditioning signal $\mathbf{c}_{\mathcal{U}}$. Finally, the generator parameterized by $\theta_{\mathcal{G}}$ predicts $\hat{y}_\mathcal{G}$ conditioned on $\mathbf{c}_{\mathcal{U}}$. 
% ${\mathcal{U}}$ represents the understanding block parameterized by $\theta_{\mathcal{U}}$ that produces a conditioning signal $\mathbf{c}_{\mathcal{U}}$ encoding world knowledge, 
% and the generator parameterized by $\theta_{\mathcal{G}}$ predicts $\hat{y}_\mathcal{G}$ conditioned on $\mathbf{c}_{\mathcal{U}}$. 
In this conditioning process, the generator must be fine-tuned to effectively utilize the semantic output from the understanding block. The training cost is relatively high in connector-based video unified models compared with image unified models, as the introduction of temporal sequences greatly increases input length and computational complexity, especially with large diffusion generators.

\subsection{Overview}
\begin{figure*}[t!]
  \centering
  \includegraphics[width=\linewidth]{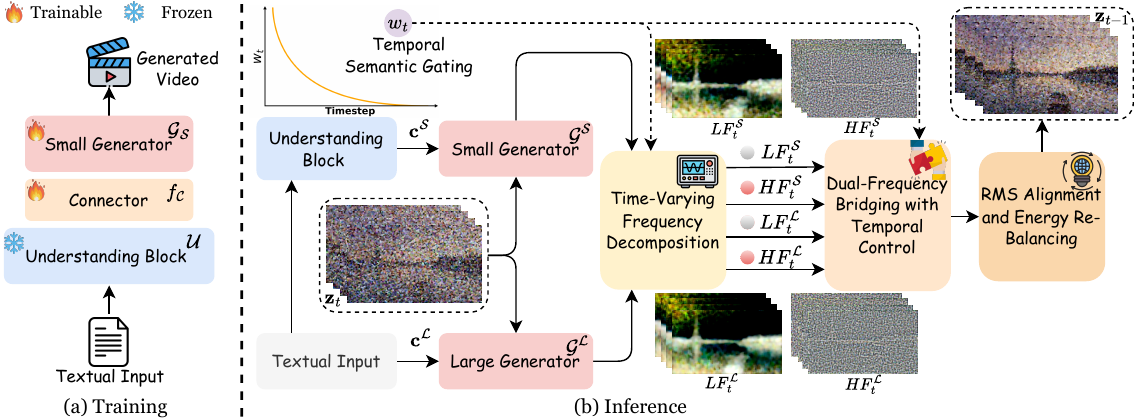}
  \caption{Overview of \ourbridge. (a): The connector and small generator are fine-tuned within the connector-based video unified model during training. (b): inference performs Unified Progressive Frequency Bridging (UPFB) to combine the small generator's semantic guidance with high-fidelity details from the large generator for high-quality video generation.}
  \vspace{-15pt}
  \label{fig:main}
\end{figure*}

% While integrating a large generator into the end-to-end understanding-to-generation loop is computationally expensive \hangjie{due to large-scale fine-tuning?}, its ability to produce high-fidelity visual details remains highly appealing. 
{While integrating a large generator into the end-to-end understanding-to-generation pipeline is computationally expensive due to the need for large-scale fine-tuning, its capacity to produce high-fidelity visual details remains highly appealing.}
To balance computational cost and generation quality, we propose \ourbridge, which leverages small and large generators that share a homogeneous latent space within connector-based unified video models, as shown in Fig.~\ref{fig:main}.
During training, only the small generator is used and fine-tuned within the unified model framework, enabling it to effectively learn to incorporate the semantic knowledge from the understanding block at an acceptable cost. 
% Exploiting the homogeneous latent space shared by the small and large generators, which enables seamless bridging between them, we adopt a training-free dual-generator scheme during inference: the small generator, trained within the unified model, acts as a semantic initiator, providing low-level semantics and global layout, while the large generator, pretrained on massive conventional datasets, is invoked as a detail refiner to supply high-level textures and photorealistic fidelity. 
Building on the shared homogeneous latent space, we design the Unified Progressive Frequency Bridging (UPFB) strategy: the small generator, trained within the unified model, primarily contributes to coherent semantics and global layout, while the large pretrained generator tends to refine details, enhance visual fidelity, and strengthens the execution of reasoning-driven semantics.
This design allows the large generator to inherit unified reasoning ability learned during training while maintaining its high visual quality, achieving superior video generation performance at a fraction of the training cost.
% This design, in which the small generator serves as an initiator for the large generator, allows the latter to inherit a substantial degree of the auxiliary generation–oriented unified reasoning ability acquired during training, while simultaneously leveraging its own high visual fidelity. Together, these factors lead to superior overall video generation quality at only a fraction of the training cost.

% To reduce training cost while preserving semantic guidance from the understanding model, we adopt a smaller generator during training. The small generator is selected to be fully homogeneous with the large generator in the latent space, which makes it a natural component in the video unified model framework for transmitting structural knowledge to the large generator.
% Therefore, at inference time, we adopt a training-free dual-generator scheme: the small generator, trained within the unified model framework, acts as a semantic initiator, providing low-level semantics and global layout, while the large generator, pretrained on massive conventional datasets, is invoked as a detail refiner to supply high-level textures and photorealistic fidelity. 
% This design inherits a substantial degree of the auxiliary generation–oriented unified reasoning ability acquired during training, while simultaneously leveraging the large model’s high visual fidelity, which together lead to superior overall video generation quality at only a fraction of the training cost.

\subsection{Unified Progressive Frequency Bridging}
% 需要强调下 计算开销这里尤其是video unified model需求比较大，顺势而然 video unified model 更需要这种方式
Simply mixing or directly bridging the outputs of the two generators often results in unstable semantics, duplicated structures, and texture conflicts, caused by their heterogeneous architectures and frequency biases. 
% \hangjie{Provide some baselines.}
To overcome this limitation, we propose Unified Progressive Frequency Bridging (UPFB), which fully exploits the small generator’s access to semantic priors from the understanding block and the large generator’s capacity for high-fidelity visual synthesis. UPFB dynamically bridges the two generators across temporal and frequency domains, treating the small generator $\mathcal{G}^\mathcal{S}$ as a semantic initiator responsible for early-stage layout and global structure, while the large generator $\mathcal{G}^\mathcal{L}$ serves as a detail refiner focusing on late-stage texture enhancement. This progressive bridging enables coherent semantic-to-detail transitions without additional training and can be seamlessly integrated into existing connector-based video unified model inference pipelines. Specifically, at sampling step $t$, both generators perform classifier-free guidance (CFG) on the same intermediate latent $\mathbf{z}_t$, but with distinct conditioning modalities:
\begin{equation}
\mathbf{v}^{{\mathcal{S}}}_t
=
\mathbf{v}^{\mathcal{S}}(\mathbf{z}_t, \mathbf{c}^{\mathcal{S}}, t)
+
s
\Big[
    \mathbf{v}^{\mathcal{S}}(\mathbf{z}_t, \mathbf{c}^{\mathcal{S}}, t)
    -
    \mathbf{v}^{\mathcal{S}}(\mathbf{z}_t, \varnothing, t)
\Big],
\end{equation}
\vspace{-5pt}
\begin{equation}
  \mathbf{v}^{{\mathcal{L}}}_t
=
\mathbf{v}^{\mathcal{L}}(\mathbf{z}_t, \mathbf{c}^{\mathcal{L}}, t)
+
s
\Big[
    \mathbf{v}^{\mathcal{L}}(\mathbf{z}_t, \mathbf{c}^{\mathcal{L}}, t)
    -
    \mathbf{v}^{\mathcal{L}}(\mathbf{z}_t, \varnothing, t)
\Big],
\end{equation}
where $\mathbf{v}^{\mathcal{S}}(\cdot)$ and $\mathbf{v}^{\mathcal{L}}(\cdot)$ denote the velocity fields predicted by $\mathcal{G}^\mathcal{S}$ and  $\mathcal{G}^\mathcal{L}$, respectively. $\mathbf{c}^{\mathcal{S}}$ represents understanding-enhanced (VLM-aligned embeddings) conditioning provided to the small generator $\mathcal{G}^\mathcal{S}$, while $\mathbf{c}^{\mathcal{L}}$ denotes the {direct conditioning} (textual embeddings) used by the large
generator $\mathcal{G}^\mathcal{L}$. $\varnothing$ is the unconditional input and $s$ is the classifier-free guidance scale.

\noindent \textbf{Temporal Semantic Gating.} Early flow-matching updates require strong global semantic control, while later steps emphasize fine-detail refinement. To facilitate a smooth coarse-to-fine transition between the two generators, we introduce a monotonic temporal weighting strategy.
\begin{align}
w_t &= \frac{1}{2}\Bigl(1 + \cos\bigl(\pi (1 - \tau_t)^{\gamma_w}\bigr)\Bigr),
\end{align}
where $\tau_t = \frac{T - 1 - t}{T - 1}$, $T$ denotes the total sampling steps, and $\gamma_w$ controls the handoff sharpness. We explicitly design the temporal weight $w_t$ to govern the dominance transition between the two generators: a larger $w_t$ biases the update
toward the small generator for semantic construction, whereas a smaller $w_t$
gradually shifts the contribution to the large generator for detail refinement.
% When $w_t \!\approx\! 1$(early stage), the small generator dominates semantic guidance; when $w_t \!\approx\! 0$ (late stage), the large generator progressively
% contributes more high-frequency details. This coarse-to-fine orchestration preserves unified reasoning ability while improving photorealistic fidelity during generation.

\noindent\textbf{Time-Varying Frequency Decomposition.}
To mitigate the frequency bias between the two generators and achieve
stable semantic-texture fusion, we explicitly separate low- and
high-frequency components in the velocity domain. Specifically, a Gaussian low-pass operator $G_{\sigma_t}(\cdot)$ is applied to the predicted velocity fields $\mathbf{v}$ to separate global layout information from fine-grained details:
\begin{equation}
\label{LF_eq}
    LF(\mathbf{v}) = G_{\sigma_t}(\mathbf{v}),
\end{equation}
\vspace{-10pt}
\begin{equation}
\label{HF_eq}
HF(\mathbf{v}) = \mathbf{v} - LF(\mathbf{v}),
\end{equation}
where $LF(\mathbf{v})$ represents the low-frequency structure of the velocity field,
encoding global semantics and coarse spatial layout, and
$HF(\mathbf{v})$ retains residual high-frequency components such as edges and fine textures. The bandwidth
parameter $\sigma_t$ decays over time to encourage a coarse-to-fine
transition:
\begin{align}
\sigma_t = \sigma_{\min} + (\sigma_{\max} - \sigma_{\min})\, w_t,
\end{align}
where both $\sigma_{\max}$ and $\sigma_{\min}$ are inference
{hyperparameters} that balance semantic stability and detail
fidelity. A large $\sigma_t$ suppresses high-frequency noise in the early
denoising stage, while a small $\sigma_t$ gradually restores fine
structures for photorealistic refinement in later stages.

\noindent\textbf{Dual-Frequency Bridging with Temporal Control.}
After frequency decoupling, the two generators exhibit complementary
strengths across different frequency bands. To effectively bridge their
outputs, we perform an asymmetric fusion in the velocity domain.
The large generator $\mathcal{G}^{\mathcal{L}}$ provides reliable
high-frequency textures, while the small generator
$\mathcal{G}^{\mathcal{S}}$ focuses on maintaining semantic coherence.
Therefore, we combine their decomposed components:

\begin{equation}
LF_t = w_t\,LF_t^\mathcal{S}
      + (1-w_t)\,LF_t^\mathcal{L},
\end{equation}

\vspace{-10pt}

\begin{equation}
    HF_t =  w_t\,\gamma_{hf}HF_t^\mathcal{S} + (1-w_t)\,HF_t^\mathcal{L},
\end{equation}
\vspace{-10pt}
\begin{equation}
    \mathbf{v}_t = LF_t + HF_t,
\end{equation}
where $LF_t^{\mathcal{S}}$ and $HF_t^{\mathcal{S}}$ denote the
low- and high-frequency components of the small generator’s guided
velocity prediction $\mathbf{v}^{\mathcal{S}}_t$, obtained through the
frequency decomposition in Eq.~\ref{LF_eq} and
Eq.~\ref{HF_eq}. Similarly, $LF_t^{\mathcal{L}}$ and
$HF_t^{\mathcal{L}}$ are derived from the large generator’s velocity
field $\mathbf{v}^{\mathcal{L}}_t$ in the same manner.
The coefficient $\gamma_{hf}\!\in\![0.5,0.8]$ is used to suppress noisy
or inconsistent high-frequency components from the small generator.
A larger $w_t$ emphasizes semantic accuracy by relying more on the
small generator, while a smaller $w_t$ gradually shifts the dominance
toward the large generator for fine-detail refinement.

\noindent\textbf{RMS Alignment and Energy Re-Balancing.}
To further strengthen the bridge between the two generators and ensure
stable integration, we normalize their velocity magnitudes across
timesteps. 
This alignment mitigates potential magnitude mismatch and
stabilizes the sampling process by applying a pre-fusion adjustment
followed by post-fusion re-balancing:
\begin{equation}
\mathbf{v}^{\mathcal{L}}_t
\leftarrow
\mathbf{v}^{\mathcal{L}}_t
\cdot
\frac{\mathrm{RMS}(\mathbf{v}^{\mathcal{S}}_t)}
     {\mathrm{RMS}(\mathbf{v}^{\mathcal{L}}_t)},
\end{equation}
\begin{equation}
\mathbf{v}_t
\leftarrow
\mathbf{v}_t
\cdot
\frac{\frac{1}{2}\bigl(\mathrm{RMS}(\mathbf{v}^{\mathcal{S}}_t)
+\mathrm{RMS}(\mathbf{v}^{\mathcal{L}}_t)\bigr)}
     {\mathrm{RMS}(\mathbf{v}_t)}.
\end{equation}
Here, $\mathrm{RMS}(\cdot)$ denotes the root-mean-square magnitude,
which quantifies the overall signal energy and ensures consistent
normalization across timesteps. This energy normalization effectively
prevents over-exposure and unstable activations, keeping the fused
velocity prediction numerically stable throughout the denoising
trajectory.

%  UPFC is a fully {training-free}, {architecture-agnostic}
% inference strategy that only operates on the predicted velocity fields
% and the shared latent state. Therefore,
In summary, UPFB can be seamlessly applied to most connector-based unified video model paradigms, where a lightweight generator learns to inherit prior knowledge from the understanding block, while a high-capacity generator operates outside the training loop to refine visual details and further strengthen the execution of reasoning-driven semantics. By simply inserting our training-free UPFB into the inference process, unified models can retain strong multimodal reasoning capabilities while effectively leveraging the high-fidelity generation power of large generators with no additional training cost. This makes UPFB a practical and scalable solution for deploying high-quality video generation in real-world unified model systems.

\subsection{VR-Bench: Benchmark for Reasoning-Driven Video Generation}
\label{sec:vrbench_main}
\begin{figure}[t!]
    \centering
    \includegraphics[width=0.8\linewidth]{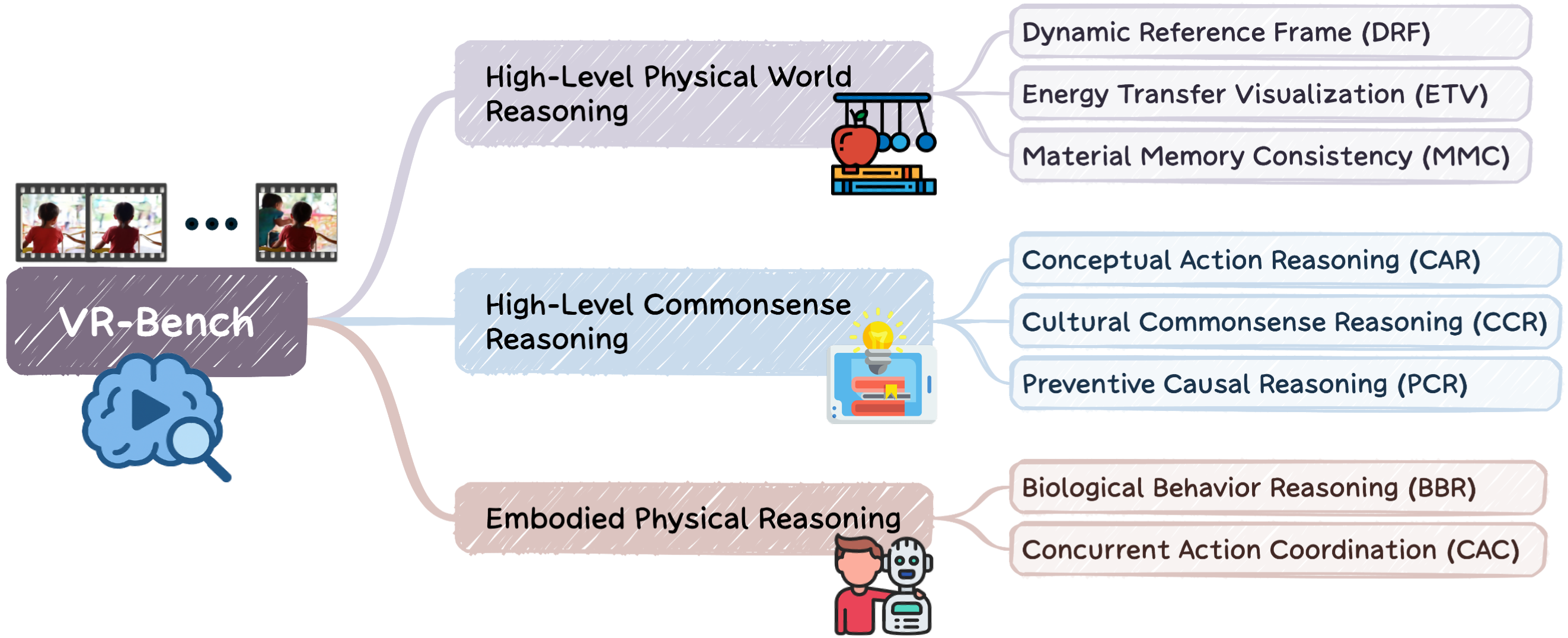}  
    \caption{Overview of VR-Bench.}
    \label{fig:example}
    \vspace{-15pt}
    \label{fig:vrbench_overview}
\end{figure}

While recent video generation benchmarks primarily focus on visual fidelity and temporal coherence, they overlook the reasoning dimension—the ability of a model to infer, plan, and act based on semantic intent. To fill this gap, we introduce \textbf{VR-Bench}, an eight-dimensional benchmark suite designed to systematically evaluate reasoning-oriented video generation. As illustrated in Fig.~\ref{fig:vrbench_overview},                    The detailed dimensions are described below.

% As illustrated in Fig.~\ref{fig:vrbench_overview}, to systematically evaluate reasoning-oriented video generation, we propose VR-Bench, a ten-dimensional benchmark suite across three reasoning categories: 
% High-Level Physical World Reasoning (capturing physical dynamics and material interactions), 
% High-Level Commonsense Reasoning (covering causal, cultural, and abstract behavioral understanding), and 
% Embodied Physical Reasoning (focusing on motion coherence and grounded physical interaction). 
% This hierarchical design provides a fine-grained diagnostic framework for analyzing reasoning capabilities across diverse video generation models.

\noindent\textbf{Dynamic Reference Frame (DRF).}
Evaluates the model’s ability to represent \textit{relative motion} and \textit{spatial relations} under changing viewpoints. Higher scores indicate improved \textit{motion coordination}, \textit{spatial consistency}, and \textit{viewpoint invariance}.
% DRF evaluates a model’s ability to represent \textit{relative motion and spatial relations} under changing viewpoints.
% It measures whether the model distinguishes absolute from relative movement in dynamic contexts.
% Higher DRF scores (0–1) indicate better \textit{motion coordination}, \textit{spatial consistency}, and \textit{viewpoint invariance}.

\noindent\textbf{Energy Transfer Visualization (ETV).}
Assesses whether generated motions reflect \textit{momentum propagation} and \textit{energy conservation}. Higher scores reflect stronger adherence to \textit{Newtonian dynamics} and \textit{temporal continuity}.
% ETV evaluates whether generated motions reflect \textit{momentum propagation} and \textit{energy conservation}.
% It measures if the model follows Newtonian principles and maintains consistent physical responses.
% Higher ETV scores (0–1) indicate better \textit{energy directionality}, \textit{conservation consistency}, and \textit{temporal continuity}.

\noindent\textbf{Material Memory Consistency (MMC).}
Measures the ability to reproduce realistic \textit{material deformation and recovery}, reflecting physical memory. Higher scores indicate more natural \textit{deformation} and \textit{relaxation dynamics}.
% MMC evaluates a model’s ability to reproduce realistic \textit{material deformation and recovery} reflecting physical memory.
% It measures whether deformable materials respond naturally to external forces and relaxation.
% Higher MMC scores (0–1) indicate better \textit{deformation realism}, \textit{recovery dynamics}, and \textit{trace consistency}.

% \noindent\textbf{Alternative Solution Adaptation (ASA).}
% Evaluates \textit{behavioral flexibility} by measuring whether the model can produce \textit{diverse yet goal-consistent} actions for the same task. Higher scores indicate better \textit{semantic consistency} and \textit{diversity under constraints}.
% ASA evaluates a model’s ability to produce \textit{diverse yet goal-consistent} behaviors for the same task (e.g., picking up a cup with either hand).
% It assesses behavioral flexibility under semantic constraints.
% Higher ASA scores (0–1) indicate better \textit{solution plausibility}, \textit{semantic consistency}, and \textit{constrained diversity}.

\noindent\textbf{Conceptual Action Reasoning (CAR).}
Tests understanding of \textit{abstract relational actions} and \textit{coherent state transitions}. Higher scores indicate stronger \textit{intent abstraction} and \textit{action-sequence coherence}.
% CAR evaluates a model’s understanding of \textit{abstract relational actions} beyond direct instruction following.
% It measures whether the model captures coherent transitions between interaction states rather than isolated motions.
% Higher CAR scores (0–1) indicate better \textit{action sequence coherence}, \textit{intentional abstraction}, and \textit{contextual–emotional coherence}.

\noindent\textbf{Cultural Commonsense Reasoning (CCR).}
Assesses the ability to generate behaviors aligned with social and cultural context. Higher scores indicate stronger \textit{symbol understanding} and \textit{social appropriateness}.
% CCR evaluates a model’s ability to generate behaviors grounded in \textit{social and cultural context}.
% It examines whether actions follow culturally appropriate semantics, symbols, and social norms.
% Higher CCR scores (0–1) indicate better \textit{symbol understanding}, \textit{contextual alignment}, and \textit{social behavior appropriateness}.

\noindent\textbf{Preventive Causal Reasoning (PCR).}
Evaluates \textit{anticipatory causal reasoning} by determining whether the model can infer \textit{preventive relations} instead of simple event sequences. Higher scores reflect clearer \textit{causal intent} and \textit{outcome consistency}.
% PCR evaluates a model’s ability to perform \textit{anticipatory causal reasoning} with clear intent understanding.
% It measures whether the model infers preventive relationships rather than simple event sequences.
% Higher PCR scores (0–1) indicate better \textit{causal chain understanding}, \textit{preventive intention clarity}, and \textit{outcome consistency}.

\noindent\textbf{Biological Behavior Reasoning (BBR).}
Measures \textit{biomechanical plausibility} and \textit{ecological coherence} in living entities. Higher scores indicate better \textit{anatomical feasibility} and \textit{environmental adaptation}.
% BBR evaluates whether generated motions exhibit \textit{biomechanical plausibility} and \textit{ecological coherence}.
% It examines if living entities move naturally and interact appropriately with their surroundings.
% Higher BBR scores (0–1) indicate better \textit{anatomical feasibility}, \textit{environmental interaction}, and \textit{rhythmic energy adaptation}.

% \noindent\textbf{Occlusion Reasoning Integrity (ORI).}
% Tests \textit{structural and depth consistency} reasoning under partial occlusion. Higher scores indicate better \textit{object permanence}, \textit{depth coherence}, and \textit{shape continuity}.
% % ORI evaluates a model’s ability to preserve \textit{structural and depth consistency} under partial occlusion.
% % It assesses whether the model maintains implicit 3D reasoning and object permanence beyond visible surfaces.
% % Higher ORI scores (0–1) indicate better \textit{occlusion ordering}, \textit{shape continuity}, and \textit{depth coherence}.

\noindent\textbf{Concurrent Action Coordination (CAC).}
Evaluates the representation of \textit{simultaneous multi-action dynamics}. Higher scores indicate stronger \textit{temporal synchronization} and \textit{semantic coherence}.
% CAC evaluates a model’s ability to represent \textit{multiple simultaneous actions} with coherent temporal and physical logic.
% It examines whether concurrent actions remain synchronized and physically feasible.
% Higher CAC scores (0–1) indicate better \textit{multi-action recognition}, \textit{temporal synchronization}, and \textit{semantic coherence}.

\section{Experiments}
\label{exp}
\subsection{Experimental Settings}
\noindent \textbf{Model.}
We adopt the connector-based video unified model Omni-Video~\cite{tan2025omni} as our baseline. The diffusion generator in Omni-Video is the Wan2.1-T2V-1.3B~\cite{wang2025wan}, which has already been fine-tuned within the unified framework. In our \ourbridge, this generation model also serves as the small generator $\mathcal{G}^\mathcal{S}$. The large generator $\mathcal{G}^\mathcal{L}$ in \ourbridge is Wan2.1-T2V-14B, which operates in the same homogeneous latent space as the small generator $\mathcal{G}^\mathcal{S}$.

\noindent \textbf{Evaluations.} 
For the text-to-video (T2V) task, we conduct a reasoning-driven evaluation using VBench~\cite{huang2024vbench} and our proposed VR-Bench. While VBench measures overall perceptual quality, VR-Bench targets reasoning alignment—assessing the consistency between inferred intent and generated content. For VR-Bench, we design 216 evaluation cases across eight dimensions, covering three high-level reasoning categories: High-Level Physical World Reasoning, High-Level Commonsense Reasoning, and Embodied Physical Reasoning. Each dimension is scored on a 0–1 scale (equivalently 0–100\%), with higher scores indicating stronger reasoning performance. All evaluations are conducted on the Qwen3-VL-30B-A3B-Instruct~\cite{yang2025qwen3} model.

\noindent \textbf{Implementation Details.}
In UPFB, we adopt a coarse-to-fine transition by jointly configuring the temporal gating sharpness $\gamma_w$ to $0.3$, the bandwidth schedule with $\sigma_{\min}=0.35$ and $\sigma_{\max}=0.70$, and the frequency-decay coefficient $\gamma_{hf}$ to $0.7$, which together provide a stable balance between semantic consistency and detail refinement. Video generation is performed at 480p resolution, with each training clip containing 81 frames (5 seconds at 16 FPS). 
During inference, we use 50 sampling steps and set the classifier-free guidance (CFG) scale to 5 for both T2I and T2V tasks.

\section{Main Results}
\subsection{Quantitative T2V Results}
We evaluate our approach on the VBench-T2V~\cite{huang2024vbench} benchmark to assess comprehensive video generation quality. As shown in Tab.~\ref{tab:vbench}, our method, \ourbridge, achieves the highest overall score of 84.12, surpassing both conventional video generation models~\cite{yuan2025lumos, hacohen2024ltx, yang2024cogvideox, kong2024hunyuanvideo, yin2025slow, wang2025wan} and recent video unified models~\cite{wang2024emu3, xie2025show, wei2025univideo, tan2025omni}. Our \ourbridge effectively integrates the strengths of both baselines, combining Omni-Video’s~\cite{tan2025omni} strong semantic understanding with Wan2.1-14B’s~\cite{wang2025wan} high-fidelity detail synthesis. This design alleviates the short-board effect, where Omni-Video’s accurate semantic guidance from the understanding block is undermined by the limited generative capacity of its smaller diffusion generator. Through unified frequency bridging, \ourbridge enhances the generator’s expressive power, achieving higher semantic alignment (79.10 → 80.52) and superior overall video generation quality.

\begin{table*}[t]
    \setlength{\tabcolsep}{2pt}
    \centering
    \definecolor{lightblue}{RGB}{240,248,255}
    \newcommand{\colorrow}[1]{\rowcolor{lightgray} #1}
    \caption{
    {Performance comparison on VBench-T2V benchmark.}
    We list partial metrics due to space limits. The \textbf{boldfacen} and \underline{underline} font indicate the highest and the second highest results.
    }
    \vspace{-5pt}
    % \vspace{-.3cm}
    \label{tab:t2v_vbench}
    \resizebox{1.\linewidth}{!}{
    \begin{tabular}{l|ccc|cccccccc}
        \toprule
        \textbf{Model}  & \textbf{Total$\uparrow$} & \textbf{Quality} & \textbf{Semantic} & \textbf{Temp. Flick.} & \textbf{Aes. Qua.} & \textbf{Obj. Class} & \textbf{Color} & \textbf{Spat. Rel.}  &  \textbf{Scene} & \textbf{App. Style} & \textbf{Overall Cons.} \\
        % \shortstack{A\\B}
        \midrule
        \multicolumn{11}{l}{\textbf{Video generation models}} \\ % Span 10 columns
        \midrule
        Lumos-1~\cite{yuan2025lumos} & 78.52 & 79.52 & 73.51 & 98.04 & 55.77 & 90.05 & 82.00 & 59.10 & 45.64 & 22.78 & 24.10 \\
        LTX-Video~\cite{hacohen2024ltx} &  80.00 & 82.30 & 70.79 & 99.34 & 59.81 & 83.45 & 81.45 & 65.43 & 51.07 & 21.47 & 25.19 \\
        CogVideoX1.5-5B~\cite{yang2024cogvideox} & 82.17 & 82.78 & \underline{79.76} & 98.88 & 62.79 & 87.47 & 87.55 & 80.25 & 52.91 & \textbf{24.89} & \underline{27.30} \\
        HunyuanVideo~\cite{kong2024hunyuanvideo} & 83.43 & 85.07 & 76.88 & 99.39 & 60.28 & 83.48 & 89.79 & 72.13 & 54.46 & 22.21 & 26.95 \\
        CausVid~\cite{yin2025slow} & \underline{83.88} & {85.21} & 78.57 & 96.89 & 64.70 & 92.80 & 80.34 & 64.77 & \underline{55.74} & 24.18 & \textbf{27.53} \\
        Wan2.1-1.3B~\cite{wang2025wan} & 83.31 & \underline{85.23} & 75.65 & 99.55 & 65.46 & 88.81 & 89.20 & 73.04 & 41.96 & 21.81 & 25.50 \\
        Wan2.1-14B~\cite{wang2025wan} & 83.69 & \textbf{85.59} & 76.11 & 99.46 & 66.07 & 86.28 & 88.59 & \underline{75.89} & 45.75 & 22.64 & 25.91 \\
        \midrule
        \multicolumn{11}{l}{\textbf{Video unified models}} \\ % Span 10 columns
        \midrule
        Emu3~\cite{wang2024emu3} & 80.96 & - & - & - & 59.64 & 86.17 & - & 68.73 & 37.11 & 20.92 & -\\         
        Show-o2~\cite{xie2025show} & 81.34 & 82.10 & 78.31 & 97.68 & 65.15 & \underline{94.81} & 80.89 & 62.61 & \textbf{57.67} & 23.29 & 27.00 \\      
        UniVideo~\cite{wei2025univideo} & 82.58 & - & - & - & - & - & - & - & - & - & - \\        
        Omni-Video~\cite{tan2025omni} & 83.82 & 85.00 & 79.10 & \underline{99.67} & \underline{66.37} & 94.56 & \underline{90.61} & 73.82 & 46.80 & 23.20 & 26.95 \\
        \midrule
        \ours & \textbf{84.12} & 85.03 & \textbf{80.52} & \textbf{99.70} & \textbf{67.29} & \textbf{96.20} & \textbf{91.46} & \textbf{76.15} & 52.47 & \underline{23.49} & 27.23 \\
        \bottomrule
    \end{tabular}}
    \vspace{-5pt}
    \label{tab:vbench}
    \end{table*}

\begin{table*}[t]
\centering
\caption{
Performance comparison on VR-Bench.
The eight metrics are grouped into three reasoning categories:
High-Level Physical World Reasoning (HL-Phys.),
High-Level Commonsense Reasoning (HL-Comm.),
and Embodied Physical Reasoning (Emb.-Phys.). Lumos-Nexus* indicates the variant constructed by replacing the large generator with Wan2.2-T2V-A14B.
\textbf{Bold} and \underline{underline} indicate the best and the second-best results, respectively.
}
\vspace{-5pt}
\label{tab:reasoning_results}
\resizebox{\linewidth}{!}{
\begin{tabular}{l|cccc|cccccccc}
\toprule
\textbf{Model} 
& \textbf{Total$\uparrow$} 
& \textbf{HL-Phys.} 
& \textbf{HL-Comm.} 
& \textbf{Emb.-Phys.} 
& \textbf{DRF} 
& \textbf{ETV} 
& \textbf{MMC} 
& \textbf{CAR} 
& \textbf{CCR} 
& \textbf{PCR} 
& \textbf{BBR} 
& \textbf{CAC} \\
 \midrule
\multicolumn{11}{l}{\textbf{Closed-source models}} \\ % Span 10 columns
\midrule
Veo 3.1~\cite{veo3.1} & {93.95} & {94.25} & {95.37} & {91.37}
& 95.80 & {93.60} & {93.33} & {97.38} & {93.52} & {95.20} & {83.85} & {98.89} \\
Kling 2.6~\cite{kling2.6} & {91.13} & {92.96} & {88.71} & {92.01}
& 99.29 & {89.60} & {90.00} & {86.26} & {91.48} & {88.40} & {88.46} & {95.56} \\
\midrule
\multicolumn{11}{l}{\textbf{Open-source models}} \\ % Span 10 columns
\midrule
CogVideoX1.5-5B~\cite{yang2024cogvideox} 
& 66.27 & 71.92 & 67.03 & 56.67 
& 93.70 & 60.40 & 61.67 & 66.82 & 76.67 & 57.60 & 69.63 & 43.70 \\
HunyuanVideo~\cite{kong2024hunyuanvideo} 
& 75.38 & 76.79 & 69.73 & \underline{81.39} 
& 94.81 & 60.00 & 75.56 & 73.47 & 78.64 & 57.07 & \textbf{79.26} & \underline{83.52} \\
Wan2.1-1.3B~\cite{wang2025wan} 
& 77.00 & 79.31 & 73.60 & 78.64 
& 95.56 & 66.13 & \underline{76.25} & 75.10 & 78.89 & 66.80 & 73.83 & {83.46} \\
Wan2.1-14B~\cite{wang2025wan} 
& \underline{78.23} & \textbf{80.34} & \underline{75.96} & 78.46 
& \underline{96.27} & \underline{66.73} & \textbf{78.03} & \textbf{77.58} & 82.96 & \underline{67.33} & 76.79 & 80.12 \\
Omni-Video~\cite{tan2025omni} 
& 72.78 & 72.39 & 72.79 & 73.33 
& 94.94 & 52.93 & 69.31 & 72.09 & \underline{83.21} & 63.07 & 74.07 & 72.59 \\ 
\ours 
& \textbf{79.28} 
& \underline{79.49} 
& \textbf{77.57} 
& \textbf{81.54} 
& \textbf{96.54} 
& \textbf{66.80} 
& 75.14 
& \underline{76.28} 
& \textbf{84.81} 
& \textbf{71.60} 
& \underline{78.89} 
& \textbf{84.20} \\
\midrule
Wan2.2-5B~\cite{wang2025wan} 
& 75.35 & 79.65 & 70.66 & 75.93
& 94.81 & 60.80 & \textbf{83.33} & 76.54 & {81.85} & 53.60 & \textbf{80.74} & 71.11 \\
Wan2.2-A14B~\cite{wang2025wan} 
& 80.98 & \textbf{83.14} & 78.86 & 80.93
& \underline{97.78} & 70.40 & \underline{81.25} & \textbf{88.46} & \underline{83.70} & \underline{64.40} & 75.56 & \underline{86.30} \\
\textbf{Lumos-Nexus*}
& \textbf{81.90} & \underline{83.02} & \textbf{79.43} & \textbf{83.93}
& \textbf{97.92} & \textbf{70.43} & 80.71 & \underline{86.23} & \textbf{84.87} & \textbf{67.19} & \underline{78.87} & \textbf{88.99} \\ \bottomrule
\end{tabular}}
\vspace{-15pt}
\end{table*}

% \subsection{VR-Bench}
We assess our method on the VR-Bench to measure reasoning-oriented video generation performance. 
In addition to our approach, we benchmark some strong closed-source models (including Veo 3.1~\cite{veo3.1} and Kling 2.6~\cite{kling2.6}) as well as a broad range of open-source models under the same evaluation protocol.
As shown in Tab.~\ref{tab:reasoning_results}, within the Wan2.1-level generation models (Rows 5–10), \ourbridge achieves the highest overall score of 79.28. 
It attains the best results in HL-Comm. (77.57) and Emb.-Phys. (81.54), demonstrating superior spatial coherence and biomechanical realism. Moreover, Lumos-Nexus effectively alleviates the short-board effect observed in Omni-Video~\cite{tan2025omni}, where accurate semantic guidance from the understanding block does not always translate into correct execution in generated videos, due to the inherent limitations of the diffusion generator’s performance.  This limitation also accounts for why Omni-Video underperforms compared with conventional video generators like Wan2.1-T2V-1.3B \cite{wang2025wan} across multiple dimensions ({\textit{e.g.}}, DRF, ETV, and CAC). In contrast, our method consistently delivers superior performance on these metrics.
To further demonstrate the extensibility of our framework and evaluate its effectiveness on stronger open-source video generation models in VR-Bench, we conduct additional experiments on the Wan2.2 series (Rows 11–13). Since Wan2.2-T2V-A14B~\cite{wang2025wan} shares a homogeneous latent space with the smaller generator used in Lumos-Nexus, we construct Lumos-Nexus* by replacing the large generator with Wan2.2-T2V-A14B. As shown in Tab.~\ref{tab:reasoning_results}, Lumos-Nexus* achieves an overall score of 81.90, outperforming Wan2.2-TI2V-5B (75.35) and Wan2.2-T2V-A14B (80.98). It further improves HL-Comm. to 79.43 and Emb.-Phys. to 83.93, while consistently achieving superior scores across multiple metrics (\textit{e.g.}, DRF 97.92, ETV 70.43, CCR 84.87, CAC 88.99). These results demonstrate that our framework improves reasoning performance and generalizes well across different video generation architectures.
\begin{figure*}[t!]
  \centering
  \includegraphics[width=\linewidth]{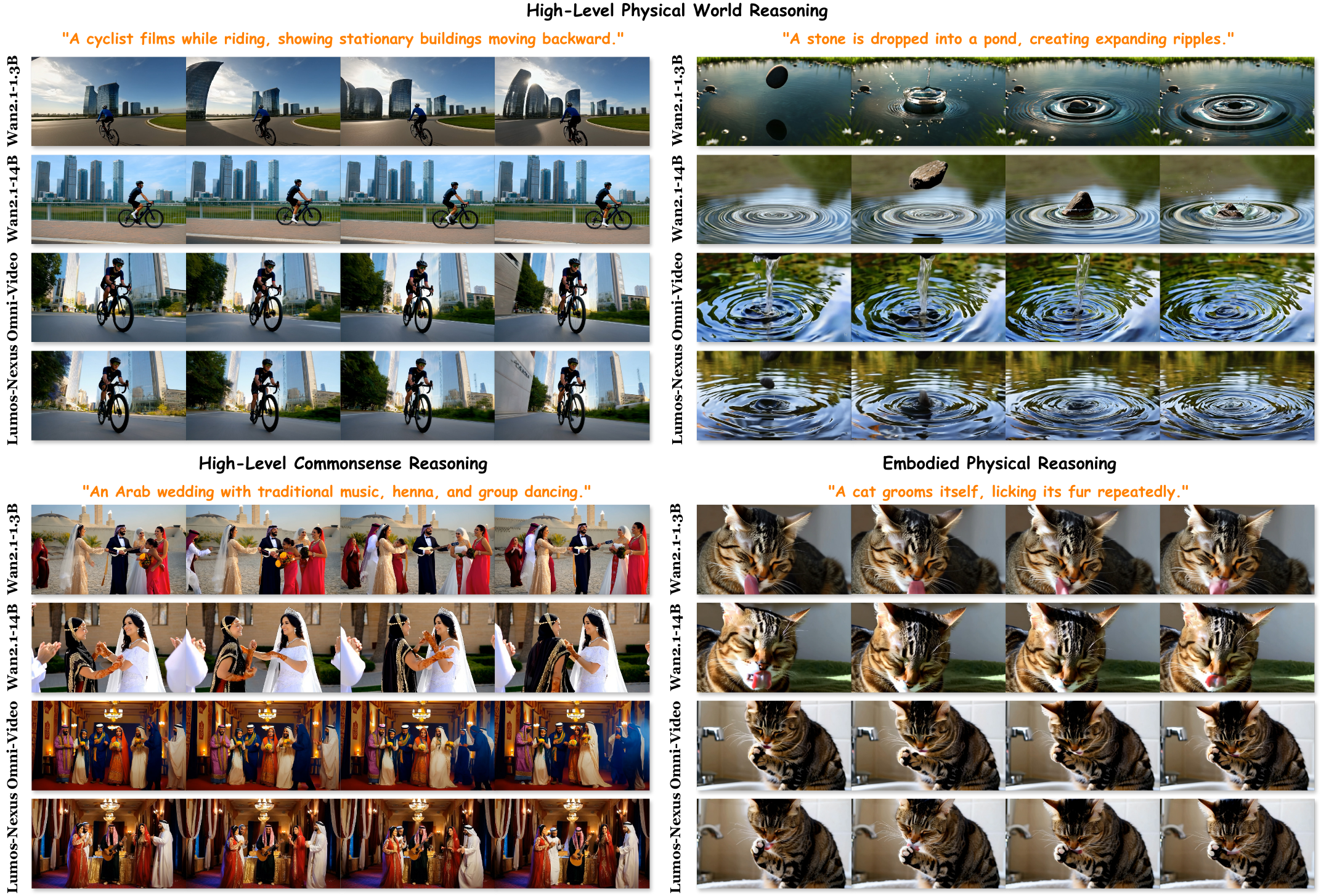}
  \caption{VR-Bench T2V qualitative comparison across three reasoning dimensions: High-Level Physical World Reasoning, High-Level Commonsense Reasoning, and Embodied Physical Reasoning.}
  \label{fig:vis_vrbench}
  \vspace{-10pt}
\end{figure*}
\subsection{Qualitative T2V Comparison}
We conduct qualitative comparisons on the three reasoning dimensions of VR-Bench, as illustrated in Fig.\ref{fig:vis_vrbench}. For High-Level Physical World Reasoning, Wan-T2V~\cite{wang2025wan} models exhibit limited ability to capture fine-grained physical regularities, such as the relative motion of stationary buildings when filming from a moving bicycle (top-left case) or the ripple propagation after a stone drops into water (top-right case). Although Omni-Video~\cite{tan2025omni} demonstrates partial awareness of such physical dynamics, its visual fidelity is limited by the capacity of its diffusion generator, even though its understanding block may provide accurate semantic priors. In contrast, our \ourbridge produces significantly more realistic and physically consistent results, adopting the same semantic priors from the understanding block as Omni-Video. Similar observations are found in High-Level Commonsense Reasoning and Embodied Physical Reasoning. Omni-Video struggles to express nuanced semantics like musical rhythm in the wedding scene (bottom-left case) and accurate tongue–paw contact in the cat grooming example (bottom-right case). Overall, \ourbridge effectively bridges the gap between Wan-2.1-14B’s limited reasoning capability and Omni-Video’s constrained generative capacity, achieving balanced improvements in high-quality video synthesis.

\begin{table*}[t!]
\caption{Performance comparison with different $\gamma_w$ on $w_t$.}
\vspace{-5pt}
\centering
\resizebox{0.8\linewidth}{!}{
\begin{tabular}{lcccccccc}
\toprule
\textbf{Method} & \multicolumn{3}{c}{\textbf{VBench}} & \multicolumn{4}{c}{\textbf{VR-Bench}} \\
\cmidrule(lr){2-4} \cmidrule(lr){5-8}
 & \textbf{Total$\uparrow$} & \textbf{Quality} & \textbf{Semantic} 
 & \textbf{Total$\uparrow$} & \textbf{HL-Phys.} & \textbf{HL-Comm.} & \textbf{Emb-Phys.} \\
\midrule
$\gamma_w=0.2$ & 84.08  &  85.09 & 80.04 & 79.09 & \textbf{80.11}  & 76.57  & 79.63  \\
$\gamma_w=0.3$ & \textbf{84.12} & 85.03 & \textbf{80.52} & \textbf{79.28} & {79.49}  & \textbf{77.57} & \textbf{81.54} \\
$\gamma_w=0.4$ & 84.02 & 85.12 & 79.63  & 76.49 & 75.77 & 75.70 & {78.77} \\
$\gamma_w=0.5$ & 84.05 & \textbf{85.24} & 79.29  & 75.10 & 73.07 & 75.89 & 76.98 \\
\bottomrule
\end{tabular}}
\label{tab:gamma_w}
    \vspace{-5pt}

\end{table*}

\section{Ablation Study}
\subsection{Impact of varying $\gamma_w$ on $w_t$}
\begin{figure*}[t!]
  \centering
  \includegraphics[width=0.85\linewidth]{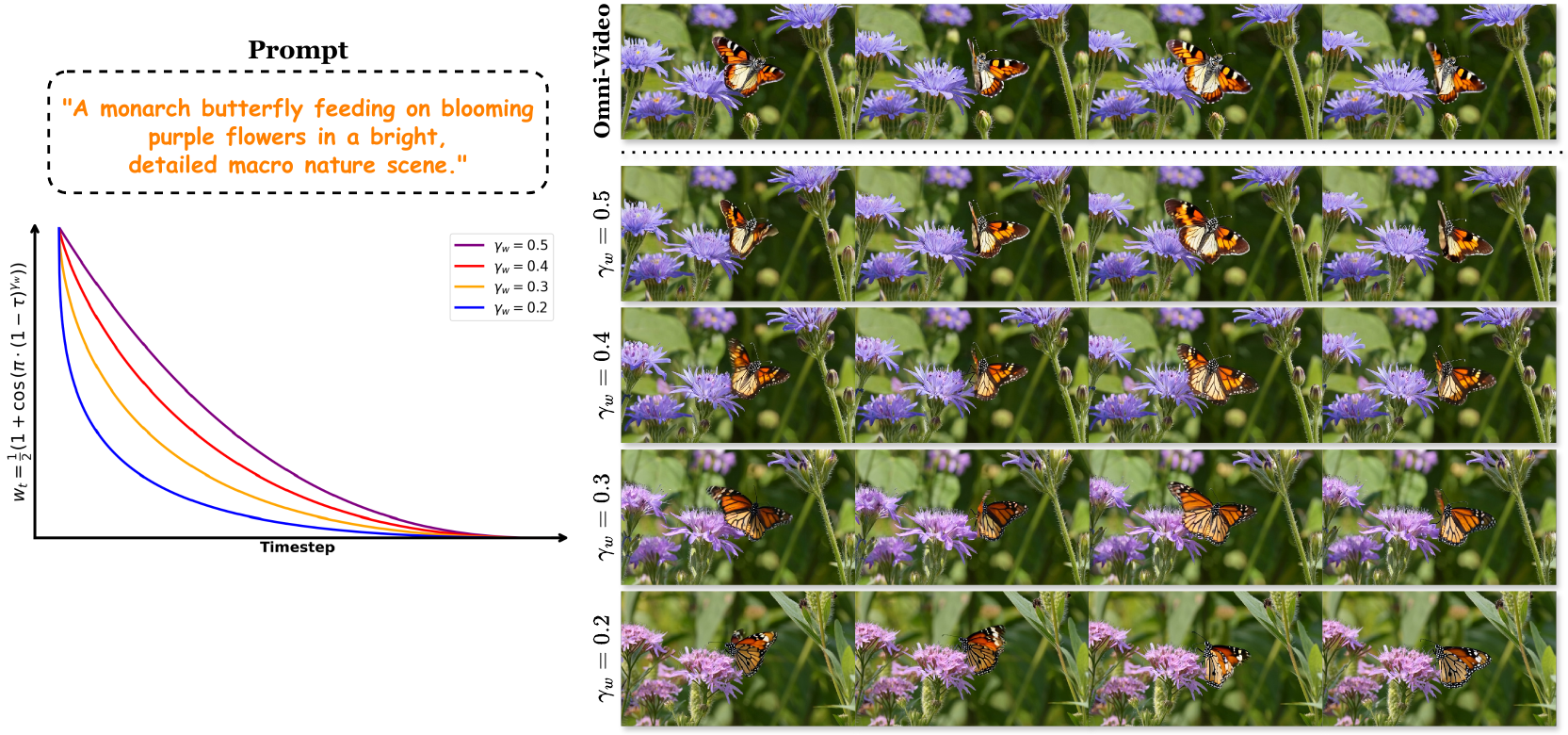}
  \vspace{-5pt}
  \caption{Visualized qualitative comparison under varying $\gamma_w$ on $w_t$.}
  \label{fig:abl_gmma_t}
  \vspace{-10pt}
\end{figure*}

We ablate the effect of the temporal transition sharpness $\gamma_w$ in UPFB, as shown in Tab.~\ref{tab:gamma_w}. As illustrated in Fig.~\ref{fig:abl_gmma_t}, $\gamma_w$ critically controls how quickly the small generator hands off semantic dominance to the large generator: when $\gamma_w$ is larger (\textit{e.g.}, $\gamma_w = 0.5$), the small generator remains influential for a longer portion of the sampling trajectory, causing the generated video to inherit more of Omni-Video’s coarse layout and camera motion patterns; conversely, smaller $\gamma_w$ values (\textit{e.g.}, $\gamma_w = 0.2$) accelerate the transition toward the large generator, leading to richer textures, stronger high-frequency details, and more vivid object appearance. Quantitatively, a moderate value ($\gamma_w$ = 0.3) yields the best overall performance across both VBench and VR-Bench: on VBench, $\gamma_w$ = 0.3 achieves the highest total score (84.12) and the strongest semantic alignment (80.52), indicating that an appropriately smooth handoff between generators benefits coarse-to-fine semantic grounding; on VR-Bench, $\gamma_w$ = 0.3 also delivers the best overall reasoning performance (79.28), particularly improving High-Level Commonsense reasoning (77.57) and Embodied Physical Reasoning (81.54). Overall, this progression shows that $\gamma_w$ effectively modulates the semantic-to-detail transition—larger values preserve baseline structural behavior, whereas smaller values enable stronger high-fidelity refinement—while extremely small or large $\gamma_w$ values weaken the balance between semantic guidance and detail refinement, confirming the necessity of a controlled, progressive transition.
% We ablate the effect of the temporal transition sharpness $\gamma_w$ in UPFB, as shown in Tab.~\ref{tab:gamma_w}. A moderate value ($\gamma_w$ = 0.3) yields the best overall performance across both VBench and VR-Bench. On VBench, $\gamma_w$ = 0.3 achieves the highest total score (84.12) and the strongest semantic alignment (80.52), indicating that an appropriately smooth handoff between generators benefits coarse-to-fine semantic grounding. On VR-Bench, $\gamma_w$ = 0.3 also delivers the best overall reasoning performance (79.28), particularly improving High-Level Commonsense reasoning (77.57) and Embodied Physical Reasoning (81.54). Extremely small or large  $\gamma_w$ values weaken the balance between semantic guidance and detail refinement, confirming the necessity of a controlled, progressive transition.

\subsection{Effect of $\sigma_{\min}$ and $\sigma_{\max}$ in UPFB}
We investigate how different bandwidth schedules $(\sigma_{\min}, \sigma_{\max})$ influence the coarse-to-fine transition of UPFB. As shown in Tab.~\ref{tab:sigmma}, a moderate setting $(\sigma_{\min}=0.35, \sigma_{\max}=0.70)$ achieves the best overall performance on VBench, yielding the highest total score (84.12). This indicates that a balanced frequency range helps maintain stable semantic grounding while progressively introducing high-frequency details. When the bandwidth is too narrow 
(0.05, 0.10) or too wide (1.00, 2.00), performance drops in both semantic and perceptual metrics.
Overly small values restrict the semantic foundation of the generation process, while excessively large values introduce unstable high-frequency components, leading to weaker coherence.
% These results highlight the importance of a balanced initial bandwidth and decay interval for effective coarse-to-fine generation.

\begin{table}[t!]
\centering
\begin{minipage}[t]{0.58\linewidth}
\centering
\caption{Performance comparison with different $(\sigma_{\min}, \sigma_{\max})$ settings
in the bandwidth schedule.}
\vspace{-5pt}
\resizebox{0.9\linewidth}{!}{
\begin{tabular}{lccc}
\toprule
\textbf{Method} & \multicolumn{3}{c}{\textbf{VBench}} \\
\cmidrule(lr){2-4}
 & \textbf{Total$\uparrow$} & \textbf{Quality} & \textbf{Semantic} \\
\midrule
$\sigma_{\min}=0.05$, $\sigma_{\max}=0.10$  & 83.99 & 84.90  & 80.34  \\
$\sigma_{\min}=0.35$, $\sigma_{\max}=0.70$  & \textbf{84.12} & \textbf{85.03}  & \textbf{80.52} \\
$\sigma_{\min}=1.00$,  $\sigma_{\max}=2.00$  & 83.56  & 84.86 & 78.36 \\
\bottomrule
\end{tabular}}
\label{tab:sigmma}
\vspace{-5pt}
\end{minipage}
\hfill
\begin{minipage}[t]{0.4\linewidth}
\centering
\caption{Performance comparison with and without RMS alignment
and energy re-balancing in UPFB.}
\vspace{-5pt}
\resizebox{\linewidth}{!}{
\begin{tabular}{lccc}
\toprule
\textbf{Method} & \multicolumn{3}{c}{\textbf{VBench}} \\
\cmidrule(lr){2-4}
 & \textbf{Total$\uparrow$} & \textbf{Quality} & \textbf{Semantic} \\
\midrule
w/o RMS  & 84.07 & 84.98  & 80.43  \\
w/ RMS  & \textbf{84.12} & \textbf{85.03}  & \textbf{80.52}  \\
\bottomrule
\end{tabular}}
\label{tab:rms}
\vspace{-5pt}
\end{minipage}

\end{table}
% \begin{table}[t!]
% \caption{Performance comparison with different $(\sigma_{\min}, \sigma_{\max})$ settings in the bandwidth schedule using VBench.}
%     \vspace{-5pt}

% \centering
% \resizebox{0.55\linewidth}{!}{
% \begin{tabular}{lccc}
% \toprule
% \textbf{Method} & \multicolumn{3}{c}{\textbf{VBench}} \\
% \cmidrule(lr){2-4}
%  & \textbf{Total} & \textbf{Quality} & \textbf{Semantic} \\
% \midrule
% $\sigma_{\min}=0.05$, $\sigma_{\max}=0.10$  & 83.99 & 84.90  & {80.34}  \\
% $\sigma_{\min}=0.35$, $\sigma_{\max}=0.70$  & \textbf{84.12} & \textbf{85.03}  & \textbf{80.52} \\
% $\sigma_{\min}=1.00$,  $\sigma_{\max}=2.00$  & 83.56  &  84.86 & 78.36 \\
% \bottomrule
% \end{tabular}}
% \label{tab:sigmma}
%     \vspace{-5pt}

% \end{table}

% \begin{table}[t!]
% \caption{Performance comparison with and without RMS alignment and energy re-balancing in UPFB.}
%     \vspace{-5pt}

% \centering
% \resizebox{0.6\linewidth}{!}{
% \begin{tabular}{lccc}
% \toprule
% \textbf{Method} & \multicolumn{3}{c}{\textbf{VBench}} \\
% \cmidrule(lr){2-4}
%  & \textbf{Total Score} & \textbf{Quality Score} & \textbf{Semantic Score} \\
% \midrule
% w/o RMS  & 84.07 & 84.98  & {80.43}  \\
% w/ RMS  & \textbf{84.12 }& \textbf{85.03}  & \textbf{{80.52}}  \\
% \bottomrule
% \end{tabular}}
% \label{tab:rms}
%     \vspace{-5pt}

% \end{table}

\subsection{Ablation of RMS Alignment and Energy Re-Balancing in UPFB}
\label{rms_energy}

To evaluate the contribution of RMS alignment and energy re-balancing in UPFB, we compare performance with and without this component, as shown in Tab.~\ref{tab:rms}. Removing RMS normalization slightly destabilizes the fusion of velocity fields from the two generators, leading to degraded semantic grounding and visual coherence. With RMS alignment enabled, UPFB achieves consistent improvements across all VBench metrics, increasing the total score from 84.07 to \textbf{84.12}, the quality score from 84.98 to {85.03}, and the semantic score from 80.43 to {80.52}. These gains confirm that RMS alignment provides a more stable integration of multi-frequency components by preventing magnitude mismatch and ensuring smoother denoising dynamics.

% \begin{table}[t!]
% \caption{Performance comparison with different $\gamma_w$ on $w_t$.}
% \centering
% \resizebox{0.9\linewidth}{!}{
% \begin{tabular}{lcccc}
% \toprule
% \textbf{Method} & \multicolumn{3}{c}{\textbf{VBench}} & \textbf{VR-Bench} \\
% \cmidrule(lr){2-4}
%  & \textbf{Total} & \textbf{Quality} & \textbf{Semantic} & \\ 
% \midrule
% $\gamma_w=0.2$ &  &  &  &72.56  \\
% $\gamma_w=0.3$ &  &  &  &72.82  \\
% $\gamma_w=0.4$ &  &  &  &69.89  \\
% $\gamma_w=0.5$ &  &  &  &69.97 \\
% \bottomrule
% \end{tabular}}
% \label{gamma_w}
% \end{table}

\section{Conclusion}
In this work, we introduce \ourbridge, a training-efficient unified video generation framework that preserves strong reasoning-driven generative capability while substantially enhancing visual fidelity. By aligning only a lightweight generator with the understanding block during training and applying Unified Progressive Frequency Bridging (UPFB) to progressively transition generation to a high-capacity pretrained generator during inference, \ourbridge enables coherent coarse-to-fine video synthesis at low training cost. To evaluate reasoning-oriented video generation, we further propose VR-Bench, an eight-dimensional benchmark that measures the alignment between inferred intent and generated content. Extensive experiments demonstrate that \ourbridge achieves significant gains in video generation.
\label{Conc}
\bibliographystyle{plain}
\bibliography{mybib}

\begin{thebibliography}{10}

\bibitem{ai2025ming}
Inclusion AI, Biao Gong, Cheng Zou, Chuanyang Zheng, Chunluan Zhou, Canxiang Yan, Chunxiang Jin, Chunjie Shen, Dandan Zheng, Fudong Wang, et~al.
\newblock Ming-omni: A unified multimodal model for perception and generation.
\newblock {\em arXiv preprint arXiv:2506.09344}, 2025.

\bibitem{blattmann2023stable}
Andreas Blattmann, Tim Dockhorn, Sumith Kulal, Daniel Mendelevitch, Maciej Kilian, Dominik Lorenz, Yam Levi, Zion English, Vikram Voleti, Adam Letts, et~al.
\newblock Stable video diffusion: Scaling latent video diffusion models to large datasets.
\newblock {\em arXiv preprint arXiv:2311.15127}, 2023.

\bibitem{chen2024videocrafter2}
Haoxin Chen, Yong Zhang, Xiaodong Cun, Menghan Xia, Xintao Wang, Chao Weng, and Ying Shan.
\newblock Videocrafter2: Overcoming data limitations for high-quality video diffusion models.
\newblock In {\em Proceedings of the IEEE/CVF Conference on Computer Vision and Pattern Recognition}, pages 7310--7320, 2024.

\bibitem{chen2025blip3}
Jiuhai Chen, Zhiyang Xu, Xichen Pan, Yushi Hu, Can Qin, Tom Goldstein, Lifu Huang, Tianyi Zhou, Saining Xie, Silvio Savarese, et~al.
\newblock Blip3-o: A family of fully open unified multimodal models-architecture, training and dataset.
\newblock {\em arXiv preprint arXiv:2505.09568}, 2025.

\bibitem{chen2025univid}
Lan Chen, Yuchao Gu, and Qi~Mao.
\newblock Univid: Unifying vision tasks with pre-trained video generation models.
\newblock {\em arXiv preprint arXiv:2509.21760}, 2025.

\bibitem{chen2025janus}
Xiaokang Chen, Zhiyu Wu, Xingchao Liu, Zizheng Pan, Wen Liu, Zhenda Xie, Xingkai Yu, and Chong Ruan.
\newblock Janus-pro: Unified multimodal understanding and generation with data and model scaling.
\newblock {\em arXiv preprint arXiv:2501.17811}, 2025.

\bibitem{deng2025emerging}
Chaorui Deng, Deyao Zhu, Kunchang Li, Chenhui Gou, Feng Li, Zeyu Wang, Shu Zhong, Weihao Yu, Xiaonan Nie, Ziang Song, et~al.
\newblock Emerging properties in unified multimodal pretraining.
\newblock {\em arXiv preprint arXiv:2505.14683}, 2025.

\bibitem{deng2024autoregressive}
Haoge Deng, Ting Pan, Haiwen Diao, Zhengxiong Luo, Yufeng Cui, Huchuan Lu, Shiguang Shan, Yonggang Qi, and Xinlong Wang.
\newblock Autoregressive video generation without vector quantization.
\newblock {\em arXiv preprint arXiv:2412.14169}, 2024.

\bibitem{dong2023dreamllm}
Runpei Dong, Chunrui Han, Yuang Peng, Zekun Qi, Zheng Ge, Jinrong Yang, Liang Zhao, Jianjian Sun, Hongyu Zhou, Haoran Wei, et~al.
\newblock Dreamllm: Synergistic multimodal comprehension and creation.
\newblock {\em arXiv preprint arXiv:2309.11499}, 2023.

\bibitem{2024SD3}
Patrick Esser, Sumith Kulal, Andreas Blattmann, Rahim Entezari, Jonas M{\"u}ller, Harry Saini, Yam Levi, Dominik Lorenz, Axel Sauer, Frederic Boesel, et~al.
\newblock Scaling rectified flow transformers for high-resolution image synthesis.
\newblock 2024.

\bibitem{ge2024seed}
Yuying Ge, Sijie Zhao, Jinguo Zhu, Yixiao Ge, Kun Yi, Lin Song, Chen Li, Xiaohan Ding, and Ying Shan.
\newblock Seed-x: Multimodal models with unified multi-granularity comprehension and generation.
\newblock {\em arXiv preprint arXiv:2404.14396}, 2024.

\bibitem{ge2024seed-x}
Yuying Ge, Sijie Zhao, Jinguo Zhu, Yixiao Ge, Kun Yi, Lin Song, Chen Li, Xiaohan Ding, and Ying Shan.
\newblock Seed-x: Multimodal models with unified multi-granularity comprehension and generation.
\newblock {\em arXiv preprint arXiv:2404.14396}, 2024.

\bibitem{ghosh2023geneval}
Dhruba Ghosh, Hannaneh Hajishirzi, and Ludwig Schmidt.
\newblock Geneval: An object-focused framework for evaluating text-to-image alignment.
\newblock {\em Advances in Neural Information Processing Systems}, 36:52132--52152, 2023.

\bibitem{veo3.1}
{Google DeepMind}.
\newblock Veo 3.1.
\newblock https://deepmind.google/models/veo/, 10 2025.

\bibitem{hacohen2024ltx}
Yoav HaCohen, Nisan Chiprut, Benny Brazowski, Daniel Shalem, Dudu Moshe, Eitan Richardson, Eran Levin, Guy Shiran, Nir Zabari, Ori Gordon, et~al.
\newblock Ltx-video: Realtime video latent diffusion.
\newblock {\em arXiv preprint arXiv:2501.00103}, 2024.

\bibitem{hong2022cogvideo}
Wenyi Hong, Ming Ding, Wendi Zheng, Xinghan Liu, and Jie Tang.
\newblock Cogvideo: Large-scale pretraining for text-to-video generation via transformers.
\newblock {\em arXiv preprint arXiv:2205.15868}, 2022.

\bibitem{huang2024vbench}
Ziqi Huang, Yinan He, Jiashuo Yu, Fan Zhang, Chenyang Si, Yuming Jiang, Yuanhan Zhang, Tianxing Wu, Qingyang Jin, Nattapol Chanpaisit, et~al.
\newblock Vbench: Comprehensive benchmark suite for video generative models.
\newblock In {\em Proceedings of the IEEE/CVF Conference on Computer Vision and Pattern Recognition}, pages 21807--21818, 2024.

\bibitem{jin2024pyramidal}
Yang Jin, Zhicheng Sun, Ningyuan Li, Kun Xu, Hao Jiang, Nan Zhuang, Quzhe Huang, Yang Song, Yadong Mu, and Zhouchen Lin.
\newblock Pyramidal flow matching for efficient video generative modeling.
\newblock {\em arXiv preprint arXiv:2410.05954}, 2024.

\bibitem{kendall1938new}
Maurice~G Kendall.
\newblock A new measure of rank correlation.
\newblock {\em Biometrika}, 30(1-2):81--93, 1938.

\bibitem{khachatryan2023text2video}
Levon Khachatryan, Andranik Movsisyan, Vahram Tadevosyan, Roberto Henschel, Zhangyang Wang, Shant Navasardyan, and Humphrey Shi.
\newblock Text2video-zero: Text-to-image diffusion models are zero-shot video generators.
\newblock In {\em Proceedings of the IEEE/CVF International Conference on Computer Vision}, pages 15954--15964, 2023.

\bibitem{kondratyuk2023videopoet}
Dan Kondratyuk, Lijun Yu, Xiuye Gu, Jos{\'e} Lezama, Jonathan Huang, Grant Schindler, Rachel Hornung, Vighnesh Birodkar, Jimmy Yan, Ming-Chang Chiu, et~al.
\newblock Videopoet: A large language model for zero-shot video generation.
\newblock {\em arXiv preprint arXiv:2312.14125}, 2023.

\bibitem{kong2024hunyuanvideo}
Weijie Kong, Qi~Tian, Zijian Zhang, Rox Min, Zuozhuo Dai, Jin Zhou, Jiangfeng Xiong, Xin Li, Bo~Wu, Jianwei Zhang, et~al.
\newblock Hunyuanvideo: A systematic framework for large video generative models.
\newblock {\em arXiv preprint arXiv:2412.03603}, 2024.

\bibitem{kling2.6}
{Kuaishou}.
\newblock Kling 2.6.
\newblock https://www.kling26.com/, 12 2025.

\bibitem{flux2024}
Black~Forest Labs.
\newblock Flux.
\newblock \url{https://github.com/black-forest-labs/flux}, 2024.

\bibitem{li2024arlon}
Zongyi Li, Shujie Hu, Shujie Liu, Long Zhou, Jeongsoo Choi, Lingwei Meng, Xun Guo, Jinyu Li, Hefei Ling, and Furu Wei.
\newblock Arlon: Boosting diffusion transformers with autoregressive models for long video generation.
\newblock {\em arXiv preprint arXiv:2410.20502}, 2024.

\bibitem{liu2024LWM}
Hao Liu, Wilson Yan, Matei Zaharia, and Pieter Abbeel.
\newblock World model on million-length video and language with blockwise ringattention.
\newblock {\em arXiv preprint arXiv:2402.08268}, 2024.

\bibitem{liu2024mardini}
Haozhe Liu, Shikun Liu, Zijian Zhou, Mengmeng Xu, Yanping Xie, Xiao Han, Juan~C P{\'e}rez, Ding Liu, Kumara Kahatapitiya, Menglin Jia, et~al.
\newblock Mardini: Masked autoregressive diffusion for video generation at scale.
\newblock {\em arXiv preprint arXiv:2410.20280}, 2024.

\bibitem{lu2023vdt}
Haoyu Lu, Guoxing Yang, Nanyi Fei, Yuqi Huo, Zhiwu Lu, Ping Luo, and Mingyu Ding.
\newblock Vdt: General-purpose video diffusion transformers via mask modeling.
\newblock {\em arXiv preprint arXiv:2305.13311}, 2023.

\bibitem{luo2025univid}
Jiabin Luo, Junhui Lin, Zeyu Zhang, Biao Wu, Meng Fang, Ling Chen, and Hao Tang.
\newblock Univid: The open-source unified video model.
\newblock {\em arXiv preprint arXiv:2509.24200}, 2025.

\bibitem{ma2024latte}
Xin Ma, Yaohui Wang, Gengyun Jia, Xinyuan Chen, Ziwei Liu, Yuan-Fang Li, Cunjian Chen, and Yu~Qiao.
\newblock Latte: Latent diffusion transformer for video generation.
\newblock {\em arXiv preprint arXiv:2401.03048}, 2024.

\bibitem{ma2025janusflow}
Yiyang Ma, Xingchao Liu, Xiaokang Chen, Wen Liu, Chengyue Wu, Zhiyu Wu, Zizheng Pan, Zhenda Xie, Haowei Zhang, Xingkai Yu, et~al.
\newblock Janusflow: Harmonizing autoregression and rectified flow for unified multimodal understanding and generation.
\newblock In {\em Proceedings of the Computer Vision and Pattern Recognition Conference}, pages 7739--7751, 2025.

\bibitem{pan2025transfer}
Xichen Pan, Satya~Narayan Shukla, Aashu Singh, Zhuokai Zhao, Shlok~Kumar Mishra, Jialiang Wang, Zhiyang Xu, Jiuhai Chen, Kunpeng Li, Felix Juefei-Xu, et~al.
\newblock Transfer between modalities with metaqueries.
\newblock {\em arXiv preprint arXiv:2504.06256}, 2025.

\bibitem{2023SDXL}
Dustin Podell, Zion English, Kyle Lacey, Andreas Blattmann, Tim Dockhorn, Jonas M{\"u}ller, Joe Penna, and Robin Rombach.
\newblock {SDXL}: Improving latent diffusion models for high-resolution image synthesis.
\newblock In {\em ICLR}, 2024.

\bibitem{qiu2025skyreels}
Di~Qiu, Zhengcong Fei, Rui Wang, Jialin Bai, Changqian Yu, Mingyuan Fan, Guibin Chen, and Xiang Wen.
\newblock Skyreels-a1: Expressive portrait animation in video diffusion transformers.
\newblock {\em arXiv preprint arXiv:2502.10841}, 2025.

\bibitem{2022DALLE2}
Aditya Ramesh, Prafulla Dhariwal, Alex Nichol, Casey Chu, and Mark Chen.
\newblock Hierarchical text-conditional image generation with {CLIP} latents.
\newblock {\em arXiv preprint arXiv:2204.06125}, 2022.

\bibitem{rombach2022stable_diffusion}
Robin Rombach, Andreas Blattmann, Dominik Lorenz, Patrick Esser, and Bj{\"o}rn Ommer.
\newblock High-resolution image synthesis with latent diffusion models.
\newblock In {\em Proceedings of the IEEE/CVF Conference on Computer Vision and Pattern Recognition}, pages 10684--10695, 2022.

\bibitem{singer2022make}
Uriel Singer, Adam Polyak, Thomas Hayes, Xi~Yin, Jie An, Songyang Zhang, Qiyuan Hu, Harry Yang, Oron Ashual, Oran Gafni, et~al.
\newblock Make-a-video: Text-to-video generation without text-video data.
\newblock {\em arXiv preprint arXiv:2209.14792}, 2022.

\bibitem{sun2024LlamaGen}
Peize Sun, Yi~Jiang, Shoufa Chen, Shilong Zhang, Bingyue Peng, Ping Luo, and Zehuan Yuan.
\newblock Autoregressive model beats diffusion: Llama for scalable image generation.
\newblock {\em arXiv preprint arXiv:2406.06525}, 2024.

\bibitem{tan2025omni}
Zhiyu Tan, Hao Yang, Luozheng Qin, Jia Gong, Mengping Yang, and Hao Li.
\newblock Omni-video: Democratizing unified video understanding and generation.
\newblock {\em arXiv preprint arXiv:2507.06119}, 2025.

\bibitem{team2024Chameleon}
Chameleon Team.
\newblock Chameleon: Mixed-modal early-fusion foundation models.
\newblock {\em arXiv preprint arXiv:2405.09818}, 2024.

\bibitem{team2025longcat}
Meituan~LongCat Team, Xunliang Cai, Qilong Huang, Zhuoliang Kang, Hongyu Li, Shijun Liang, Liya Ma, Siyu Ren, Xiaoming Wei, Rixu Xie, et~al.
\newblock Longcat-video technical report.
\newblock {\em arXiv preprint arXiv:2510.22200}, 2025.

\bibitem{teng2025magi}
Hansi Teng, Hongyu Jia, Lei Sun, Lingzhi Li, Maolin Li, Mingqiu Tang, Shuai Han, Tianning Zhang, WQ~Zhang, Weifeng Luo, et~al.
\newblock Magi-1: Autoregressive video generation at scale.
\newblock {\em arXiv preprint arXiv:2505.13211}, 2025.

\bibitem{tong2025metamorph}
Shengbang Tong, David Fan, Jiachen Li, Yunyang Xiong, Xinlei Chen, Koustuv Sinha, Michael Rabbat, Yann LeCun, Saining Xie, and Zhuang Liu.
\newblock Metamorph: Multimodal understanding and generation via instruction tuning.
\newblock In {\em Proceedings of the IEEE/CVF International Conference on Computer Vision}, pages 17001--17012, 2025.

\bibitem{wang2025wan}
Ang Wang, Baole Ai, Bin Wen, Chaojie Mao, Chen-Wei Xie, Di~Chen, Feiwu Yu, Haiming Zhao, Jianxiao Yang, Jianyuan Zeng, et~al.
\newblock Wan: Open and advanced large-scale video generative models.
\newblock {\em arXiv preprint arXiv:2503.20314}, 2025.

\bibitem{wang2023modelscope}
Jiuniu Wang, Hangjie Yuan, Dayou Chen, Yingya Zhang, Xiang Wang, and Shiwei Zhang.
\newblock Modelscope text-to-video technical report.
\newblock {\em arXiv preprint arXiv:2308.06571}, 2023.

\bibitem{wang2024emu3}
Xinlong Wang, Xiaosong Zhang, Zhengxiong Luo, Quan Sun, Yufeng Cui, Jinsheng Wang, Fan Zhang, Yueze Wang, Zhen Li, Qiying Yu, et~al.
\newblock Emu3: Next-token prediction is all you need.
\newblock {\em arXiv preprint arXiv:2409.18869}, 2024.

\bibitem{wang2024loong}
Yuqing Wang, Tianwei Xiong, Daquan Zhou, Zhijie Lin, Yang Zhao, Bingyi Kang, Jiashi Feng, and Xihui Liu.
\newblock Loong: Generating minute-level long videos with autoregressive language models.
\newblock {\em arXiv preprint arXiv:2410.02757}, 2024.

\bibitem{wei2025univideo}
Cong Wei, Quande Liu, Zixuan Ye, Qiulin Wang, Xintao Wang, Pengfei Wan, Kun Gai, and Wenhu Chen.
\newblock Univideo: Unified understanding, generation, and editing for videos.
\newblock {\em arXiv preprint arXiv:2510.08377}, 2025.

\bibitem{wu2025janus}
Chengyue Wu, Xiaokang Chen, Zhiyu Wu, Yiyang Ma, Xingchao Liu, Zizheng Pan, Wen Liu, Zhenda Xie, Xingkai Yu, Chong Ruan, et~al.
\newblock Janus: Decoupling visual encoding for unified multimodal understanding and generation.
\newblock In {\em Proceedings of the Computer Vision and Pattern Recognition Conference}, pages 12966--12977, 2025.

\bibitem{wu2023tune}
Jay~Zhangjie Wu, Yixiao Ge, Xintao Wang, Stan~Weixian Lei, Yuchao Gu, Yufei Shi, Wynne Hsu, Ying Shan, Xiaohu Qie, and Mike~Zheng Shou.
\newblock Tune-a-video: One-shot tuning of image diffusion models for text-to-video generation.
\newblock In {\em Proceedings of the IEEE/CVF international conference on computer vision}, pages 7623--7633, 2023.

\bibitem{wu2024next}
Shengqiong Wu, Hao Fei, Leigang Qu, Wei Ji, and Tat-Seng Chua.
\newblock Next-gpt: Any-to-any multimodal llm.
\newblock In {\em Forty-first International Conference on Machine Learning}, 2024.

\bibitem{xie2024show}
Jinheng Xie, Weijia Mao, Zechen Bai, David~Junhao Zhang, Weihao Wang, Kevin~Qinghong Lin, Yuchao Gu, Zhijie Chen, Zhenheng Yang, and Mike~Zheng Shou.
\newblock Show-o: One single transformer to unify multimodal understanding and generation.
\newblock {\em arXiv preprint arXiv:2408.12528}, 2024.

\bibitem{xie2024show-o}
Jinheng Xie, Weijia Mao, Zechen Bai, David~Junhao Zhang, Weihao Wang, Kevin~Qinghong Lin, Yuchao Gu, Zhijie Chen, Zhenheng Yang, and Mike~Zheng Shou.
\newblock Show-o: One single transformer to unify multimodal understanding and generation.
\newblock {\em arXiv preprint arXiv:2408.12528}, 2024.

\bibitem{xie2025show}
Jinheng Xie, Zhenheng Yang, and Mike~Zheng Shou.
\newblock Show-o2: Improved native unified multimodal models.
\newblock {\em arXiv preprint arXiv:2506.15564}, 2025.

\bibitem{xing2026lumosx}
Jiazheng Xing, Fei Du, Hangjie Yuan, Pengwei Liu, Hongbin Xu, Hai Ci, Ruigang Niu, Weihua Chen, Fan Wang, and Yong Liu.
\newblock Lumosx: Relate any identities with their attributes for personalized video generation.
\newblock In {\em The Fourteenth International Conference on Learning Representations}, 2026.

\bibitem{yan2021videogpt}
Wilson Yan, Yunzhi Zhang, Pieter Abbeel, and Aravind Srinivas.
\newblock Videogpt: Video generation using vq-vae and transformers.
\newblock {\em arXiv preprint arXiv:2104.10157}, 2021.

\bibitem{yang2025qwen3}
An~Yang, Anfeng Li, Baosong Yang, Beichen Zhang, Binyuan Hui, Bo~Zheng, Bowen Yu, Chang Gao, Chengen Huang, Chenxu Lv, et~al.
\newblock Qwen3 technical report.
\newblock {\em arXiv preprint arXiv:2505.09388}, 2025.

\bibitem{yang2025mmada}
Ling Yang, Ye~Tian, Bowen Li, Xinchen Zhang, Ke~Shen, Yunhai Tong, and Mengdi Wang.
\newblock Mmada: Multimodal large diffusion language models.
\newblock {\em arXiv preprint arXiv:2505.15809}, 2025.

\bibitem{yang2024cogvideox}
Zhuoyi Yang, Jiayan Teng, Wendi Zheng, Ming Ding, Shiyu Huang, Jiazheng Xu, Yuanming Yang, Wenyi Hong, Xiaohan Zhang, Guanyu Feng, et~al.
\newblock Cogvideox: Text-to-video diffusion models with an expert transformer.
\newblock {\em arXiv preprint arXiv:2408.06072}, 2024.

\bibitem{yin2025slow}
Tianwei Yin, Qiang Zhang, Richard Zhang, William~T Freeman, Fredo Durand, Eli Shechtman, and Xun Huang.
\newblock From slow bidirectional to fast autoregressive video diffusion models.
\newblock In {\em Proceedings of the Computer Vision and Pattern Recognition Conference}, pages 22963--22974, 2025.

\bibitem{yuan2025lumos}
Hangjie Yuan, Weihua Chen, Jun Cen, Hu~Yu, Jingyun Liang, Shuning Chang, Zhihui Lin, Tao Feng, Pengwei Liu, Jiazheng Xing, et~al.
\newblock Lumos-1: On autoregressive video generation from a unified model perspective.
\newblock {\em arXiv preprint arXiv:2507.08801}, 2025.

\bibitem{zhang2025show}
David~Junhao Zhang, Jay~Zhangjie Wu, Jia-Wei Liu, Rui Zhao, Lingmin Ran, Yuchao Gu, Difei Gao, and Mike~Zheng Shou.
\newblock Show-1: Marrying pixel and latent diffusion models for text-to-video generation.
\newblock {\em International Journal of Computer Vision}, 133(4):1879--1893, 2025.

\bibitem{zheng2024open}
Zangwei Zheng, Xiangyu Peng, Tianji Yang, Chenhui Shen, Shenggui Li, Hongxin Liu, Yukun Zhou, Tianyi Li, and Yang You.
\newblock Open-sora: Democratizing efficient video production for all.
\newblock {\em arXiv preprint arXiv:2412.20404}, 2024.

\bibitem{zhou2024transfusion}
Chunting Zhou, Lili Yu, Arun Babu, Kushal Tirumala, Michihiro Yasunaga, Leonid Shamis, Jacob Kahn, Xuezhe Ma, Luke Zettlemoyer, and Omer Levy.
\newblock Transfusion: Predict the next token and diffuse images with one multi-modal model.
\newblock {\em arXiv preprint arXiv:2408.11039}, 2024.

\end{thebibliography}

\onecolumn
\appendix
\begin{center}
{\Large\bfseries Appendix of \ours \par}
\end{center}
In this Appendix, we provide additional content organized as follows:
\begin{itemize}
  \item \textbf{Sec.~\ref{alg_workflow}} presents the {algorithmic workflow of Lumos-Nexus}.
  \item \textbf{Sec.~\ref{app:vrbench}} provides {detailed descriptions of VR-Bench}.
  \item \textbf{Sec.~\ref{t2i-result}} includes additional {quantitative text-to-image (T2I) results}.
  \item \textbf{Sec.~\ref{app_ablation}} presents {more ablation discussions}, including:
    \begin{itemize}
      \item Sec.~\ref{bridge_models} Discussion on bridging large and small generators in the video unified model.  
      \item Sec.~\ref{cross_eval} VR-Bench Cross-Evaluator Robustness.
      \item Sec.~\ref{diss_effi} Discussion of Model Efficiency.
      \item Sec.~\ref{UPFB_gains} Disentangling Model Capacity from UPFB Gains.
      \item Sec.~\ref{more_hyper} More Discussions of Hyperparameters.
      \item Sec.~\ref{human_eval} Human Evaluation on VR-Bench.
      \item Sec.~\ref{assump_latent_space} Discussion of the Homogeneous Latent Space Assumption.
    \item Sec.~\ref{generality} Discussion of Lumos-Nexus's Generality.
    \end{itemize}
 \item \textbf{Sec.~\ref{QAs_VR-Bench}} shows {Q\&A evaluation examples from VR-Bench}.
 \item \textbf{Sec.~\ref{limit_label}} discusses the limitations.
\end{itemize}

\section{Algorithmic Workflow of \ours}
\label{alg_workflow}
The overall procedure of Unified Progressive Frequency Bridging (UPFB), detailed in Sec.~3.3 of the main text, is summarized in Algorithm \ref{alg:upfb}. UPFB enables inference-time bridging between the lightweight semantic generator and the large high-fidelity generator through a temporally controlled, frequency-aware fusion process. At each step, both models predict velocity fields with classifier-free guidance, which are then gated by a cosine-scheduled temporal weight. A time-varying Gaussian filter decomposes the velocities into low- and high-frequency components, facilitating stable semantic–texture separation. These components are asymmetrically fused so that the small generator guides early semantic formation while the large generator gradually injects high-frequency detail. RMS normalization before and after fusion maintains consistent signal scale. The fused velocity field is finally used for latent updates via the flow-matching solver, yielding a trajectory that maintains and further strengthens semantic coherence while enhancing photorealistic fidelity.

\begin{algorithm}[t]
{\small
\caption{Unified Progressive Frequency Bridging (UPFB)}
\label{alg:upfb}
\begin{algorithmic}[1]
\Require Initial latent $\mathbf{z}_T$, total steps $T$, CFG scale $s$, 
temporal sharpness $\gamma_w$, HF suppression $\gamma_{hf}$, 
bandwidth range $(\sigma_{\min}, \sigma_{\max})$, 
conditionings $\mathbf{c}^\mathcal{S}$ (understanding-enhanced) and $\mathbf{c}^\mathcal{L}$ (direct text)
\Ensure Refined latent $\mathbf{z}_0$ for video decoding
\For{$t = T-1, \dots, 0$}
    \State \textbf{Temporal semantic gating:}
    \State $\tau_t \gets \dfrac{T-1-t}{T-1}$
    \State $w_t \gets \dfrac{1}{2}\Bigl(1 + \cos\bigl(\pi (1-\tau_t)^{\gamma_w}\bigr)\Bigr)$
    \State \textbf{Velocity prediction with CFG:}
    \State $\mathbf{v}_t^{\mathcal{S}} \gets 
        \mathbf{v}^{\mathcal{S}}(\mathbf{z}_t,\mathbf{c}^{\mathcal{S}},t)
        + s\bigl[
        \mathbf{v}^{\mathcal{S}}(\mathbf{z}_t,\mathbf{c}^{\mathcal{S}},t)
        - \mathbf{v}^{\mathcal{S}}(\mathbf{z}_t,\varnothing,t)
        \bigr]$
    \State $\mathbf{v}_t^{\mathcal{L}} \gets 
        \mathbf{v}^{\mathcal{L}}(\mathbf{z}_t,\mathbf{c}^{\mathcal{L}},t)
        + s\bigl[
        \mathbf{v}^{\mathcal{L}}(\mathbf{z}_t,\mathbf{c}^{\mathcal{L}},t)
        - \mathbf{v}^{\mathcal{L}}(\mathbf{z}_t,\varnothing,t)
        \bigr]$
    \State \textbf{RMS pre-alignment:}
    \State $\mathbf{v}_t^{\mathcal{L}} \gets \mathbf{v}_t^{\mathcal{L}}
        \cdot \dfrac{\mathrm{RMS}(\mathbf{v}_t^{\mathcal{S}})}
                     {\mathrm{RMS}(\mathbf{v}_t^{\mathcal{L}})}$
    \State \textbf{Time-varying frequency decomposition:}
    \State $\sigma_t \gets \sigma_{\min} + (\sigma_{\max}-\sigma_{\min})\,w_t$
    \State $LF_t^{\mathcal{S}} \gets G_{\sigma_t}(\mathbf{v}_t^{\mathcal{S}})$;\quad
           $HF_t^{\mathcal{S}} \gets \mathbf{v}_t^{\mathcal{S}} - LF_t^{\mathcal{S}}$
    \State $LF_t^{\mathcal{L}} \gets G_{\sigma_t}(\mathbf{v}_t^{\mathcal{L}})$;\quad
           $HF_t^{\mathcal{L}} \gets \mathbf{v}_t^{\mathcal{L}} - LF_t^{\mathcal{L}}$
    \State \textbf{Dual-frequency bridging with temporal control:}
    \State $LF_t \gets w_t\,LF_t^{\mathcal{S}} + (1-w_t)\,LF_t^{\mathcal{L}}$
    \State $HF_t \gets w_t\,\gamma_{hf}\,HF_t^{\mathcal{S}} + (1-w_t)\,HF_t^{\mathcal{L}}$
    \State $\mathbf{v}_t \gets LF_t + HF_t$
    \State \textbf{RMS re-balancing:}
    \State $\mathbf{v}_t \gets \mathbf{v}_t \cdot 
        \dfrac{\tfrac{1}{2}\bigl(\mathrm{RMS}(\mathbf{v}_t^{\mathcal{S}})
        + \mathrm{RMS}(\mathbf{v}_t^{\mathcal{L}})\bigr)}
              {\mathrm{RMS}(\mathbf{v}_t)}$
    \State \textbf{Latent update (flow-matching step):}
    \State $\mathbf{z}_{t-1} \gets \mathbf{z}_t$ updated using $\mathbf{v}_t$
\EndFor
\State \Return $\mathbf{z}_0$
\end{algorithmic}
}
\end{algorithm}

\begin{figure*}[b!]
\centering
\includegraphics[width=0.95\linewidth]{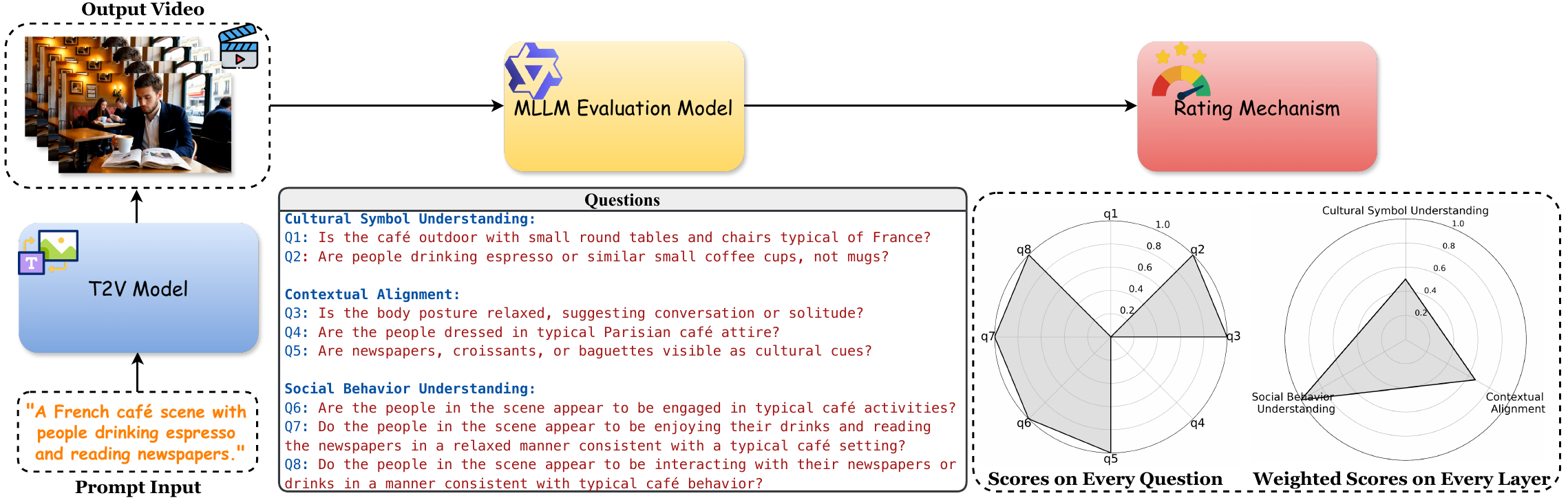}
\caption{
Evaluation pipeline for VR-Bench.
The MLLM answers metric-specific questions based on its analysis and understanding of the generated video.
}
\label{fig:eval_pipeline}
\end{figure*}

\section{Detailed Descriptions of VR-Bench}
\label{app:vrbench}

We provide detailed descriptions of the eight reasoning dimensions in \textbf{VR-Bench}, as introduced in Sec.~3.4 of the main text, which span three hierarchical categories: (1) \textit{High-Level Physical World Reasoning}, capturing physical dynamics and material interactions; (2) \textit{High-Level Commonsense Reasoning}, assessing causal, cultural, and abstract behavioral understanding; and (3) \textit{Embodied Physical Reasoning}, focusing on motion coherence and grounded physical interactions. Each dimension is evaluated using an eight-question prompt design for fine-grained assessment. As shown in Fig.~\ref{fig:eval_pipeline}, for each metric dimension, we construct a set of video-generation prompts and derive metric-specific questions for each prompt. Given a prompt, the video generation model produces output frames, which are subsequently evaluated by an MLLM (Qwen3-VL-30B-A3B-Instruct~\cite{yang2025qwen3}). The MLLM analyzes the generated content and provides binary responses (“yes’’ or “no’’) to each question. These responses are converted into numerical scores (1 for “yes’’ 0 for “no’’) and aggregated using a layer-wise weighting scheme to obtain the final metric score. 
% The detailed evaluation procedure is provided below.
% \subsection{Layered Evaluation Scheme}
% \label{app:layered}

Each metric in VR-Bench evaluates one of the eight reasoning aspects and adopts a unified eight-question structure. The questions are organized into three layers that reflect a common reasoning pattern shared across these aspects. Details are as follows:

\noindent \textbf{Layer 1 (Q1–Q2): Basic perception.}  
These questions examine fundamental visual facts, such as object identity, motion direction, contact events, or culturally meaningful symbols. The model must first recognize the essential elements of the scene.

\noindent \textbf{Layer 2 (Q3–Q5): Mid-level relational reasoning.}  
These questions evaluate interactions among objects, agents, and physical signals, including relative motion, force responses, material deformation, cultural context, and multi-agent dynamics.

\noindent \textbf{Layer 3 (Q6–Q8): High-level causal or semantic reasoning.}  
These questions probe deeper understanding, requiring the model to infer intention, causal flow, energy consistency, cultural norms, or biological plausibility across time.

Together, the three layers provide a consistent and interpretable structure for evaluating diverse forms of reasoning in video generation. Further details for each metric dimension are provided below.

% -----------------------------------------------
\subsection{High-Level Physical World Reasoning}
% -----------------------------------------------

\subsubsection{Dynamic Reference Frame (DRF)}
\label{app:drf}

\textbf{Target capability.}  
DRF evaluates whether a model preserves stable spatial relationships when the viewpoint changes. The model should align relative object motion with the corresponding camera motion.

\noindent\textbf{Motivation.}  
Moving cameras or platforms introduce parallax and geometric shifts. A robust model must correctly infer which entities are moving and adjust background motion to remain consistent with the camera trajectory.

\noindent\textbf{Evaluation aspects.}  
Scene understanding (Q1–Q2) examines viewpoint attachment and global layout.  
Relational dynamics (Q3–Q5) assess whether relative motion and parallax follow physical rules.  
Physical–temporal coherence (Q6–Q8) checks motion smoothness, micro-vibrations, and lighting stability.

\noindent\textbf{Expected behavior.}  
Spatial relations should evolve smoothly with viewpoint motion, exhibiting correct parallax and physically consistent dynamics.

% ------------------------------------------------
\subsubsection{Energy Transfer Visualization (ETV)}
\label{app:etv}

\textbf{Target capability.}  
ETV evaluates whether a model adheres to basic principles of energy transfer, including force direction, momentum propagation, and energy decay.

\noindent\textbf{Motivation.}  
Smooth trajectories alone do not guarantee physical correctness. ETV tests a model’s ability to represent how energy travels through a scene and dissipates over time.

\noindent\textbf{Evaluation aspects.}  
Energy continuity (Q1–Q2) checks immediate responses to forces or discharges.  
Directional consistency (Q3–Q5) examines rebounds, illumination changes, and decay rates.  
Physical plausibility (Q6–Q8) verifies smooth attenuation without abrupt transitions.

\noindent\textbf{Expected behavior.}  
Energy transmission and fading should follow clear, continuous physical patterns.

% ------------------------------------------------
\subsubsection{Material Memory Consistency (MMC)}
\label{app:mmc}

\textbf{Target capability.}  
MMC evaluates whether material deformation and recovery follow their inherent physical properties.

\noindent\textbf{Motivation.}  
Different materials—such as clay, rubber, or fabric—deform and revert in distinct ways. Models often fail to reproduce correct elasticity or residual marks.

\noindent\textbf{Evaluation aspects.}  
Deformation realism (Q1–Q2) checks whether applied pressure yields plausible shape changes.  
Recovery dynamics (Q3–Q5) evaluate how the material reverts and whether residual deformation remains.  
Material trace consistency (Q6–Q8) assesses texture stability, lighting coherence, and lasting deformation cues.

\noindent\textbf{Expected behavior.}  
Deformation and recovery should match the material’s physical identity, including plausible residual marks.

% =========================================================
\subsection{High-Level Commonsense Reasoning}
% =========================================================

\subsubsection{Conceptual Action Reasoning (CAR)}
\label{app:car}

\textbf{Target capability.}  
CAR evaluates whether the model represents actions with coherent intentions rather than mere motion.

\noindent\textbf{Motivation.}  
Human actions typically follow goals and roles. CAR tests whether the model can portray purposeful action sequences.

\noindent\textbf{Evaluation aspects.}  
Action sequence coherence (Q1–Q2) checks logical ordering.  
Relational dynamics (Q3–Q5) examine agent interactions and their causal influence.  
Intent and temporal coherence (Q6–Q8) evaluate whether actions align with the overarching goal.

\noindent\textbf{Expected behavior.}  
Actions should progress logically, reflecting consistent intentions across time.

% ------------------------------------------------
\subsubsection{Cultural Commonsense Reasoning (CCR)}
\label{app:ccr}

\textbf{Target capability.}  
CCR evaluates whether scenes, objects, and behaviors follow culturally grounded meanings and practices.

\noindent\textbf{Motivation.}  
Cultural understanding goes beyond visual recognition; it requires knowledge of objects, customs, and social norms.

\noindent\textbf{Evaluation aspects.}  
Cultural symbol understanding (Q1–Q2) checks correct appearance of cultural objects.  
Contextual alignment (Q3–Q5) evaluates setting, attire, and activities.  
Social behavior understanding (Q6–Q8) assesses whether interactions follow cultural expectations.

\noindent\textbf{Expected behavior.}  
Scenes should present objects, environments, and behaviors consistent with the intended cultural context.

% ------------------------------------------------
\subsubsection{Preventive Causal Reasoning (PCR)}
\label{app:pcr}

\textbf{Target capability.}  
PCR evaluates whether a model can represent proactive actions taken to prevent negative outcomes.

\noindent\textbf{Motivation.}  
Many models react only after an event unfolds. PCR tests the ability to detect early risk cues and respond preemptively.

\noindent\textbf{Evaluation aspects.}  
Causal anticipation (Q1–Q2) identifies early warnings.  
Preventive timing (Q3–Q5) evaluates whether responses occur early enough.  
Visual plausibility (Q6–Q8) checks the clarity of the cue–action–outcome chain.

\noindent\textbf{Expected behavior.}  
Early cues should trigger timely, effective actions that avert the negative event.

% =========================================================
\subsection{Embodied Physical Reasoning}
% =========================================================

\subsubsection{Biological Behavior Reasoning (BBR)}
\label{app:bbr}

\textbf{Target capability.}  
BBR evaluates whether organism motion respects anatomical constraints and responds realistically to the environment.

\noindent\textbf{Motivation.}  
Real biological movement relies on joint limits, coordination, and contact dynamics. Deviations in these aspects signal weak physical modeling.

\noindent\textbf{Evaluation aspects.}  
Biomechanical realism (Q1–Q2) checks push-off patterns, deceleration, and locomotion.  
Environmental interaction (Q3–Q5) evaluates contact responses, balance, and reflections.  
Ecological coherence (Q6–Q8) assesses posture, surface interaction, and lighting coherence.

\noindent\textbf{Expected behavior.}  
Movement should respect anatomical principles and interact consistently with the surrounding environment.

% ------------------------------------------------
\subsubsection{Concurrent Action Coordination (CAC)}
\label{app:cac}

\textbf{Target capability.}  
CAC evaluates whether a model can depict multiple simultaneous actions while maintaining internal consistency.

\noindent\textbf{Motivation.}  
Humans often perform several actions concurrently, and models may lose synchronization or drop secondary actions.

\noindent\textbf{Evaluation aspects.}  
Temporal synchronization (Q1–Q2) checks inter-action timing.  
Attention coordination (Q3–Q5) evaluates gaze alignment, posture control, and task focus.  
Physical realism (Q6–Q8) checks whether simultaneous actions are physically compatible.

\noindent\textbf{Expected behavior.}  
Concurrent actions should remain coordinated in timing, posture, and physical interaction.

\begin{figure*}[th!]
  \centering
  \includegraphics[width=\linewidth]{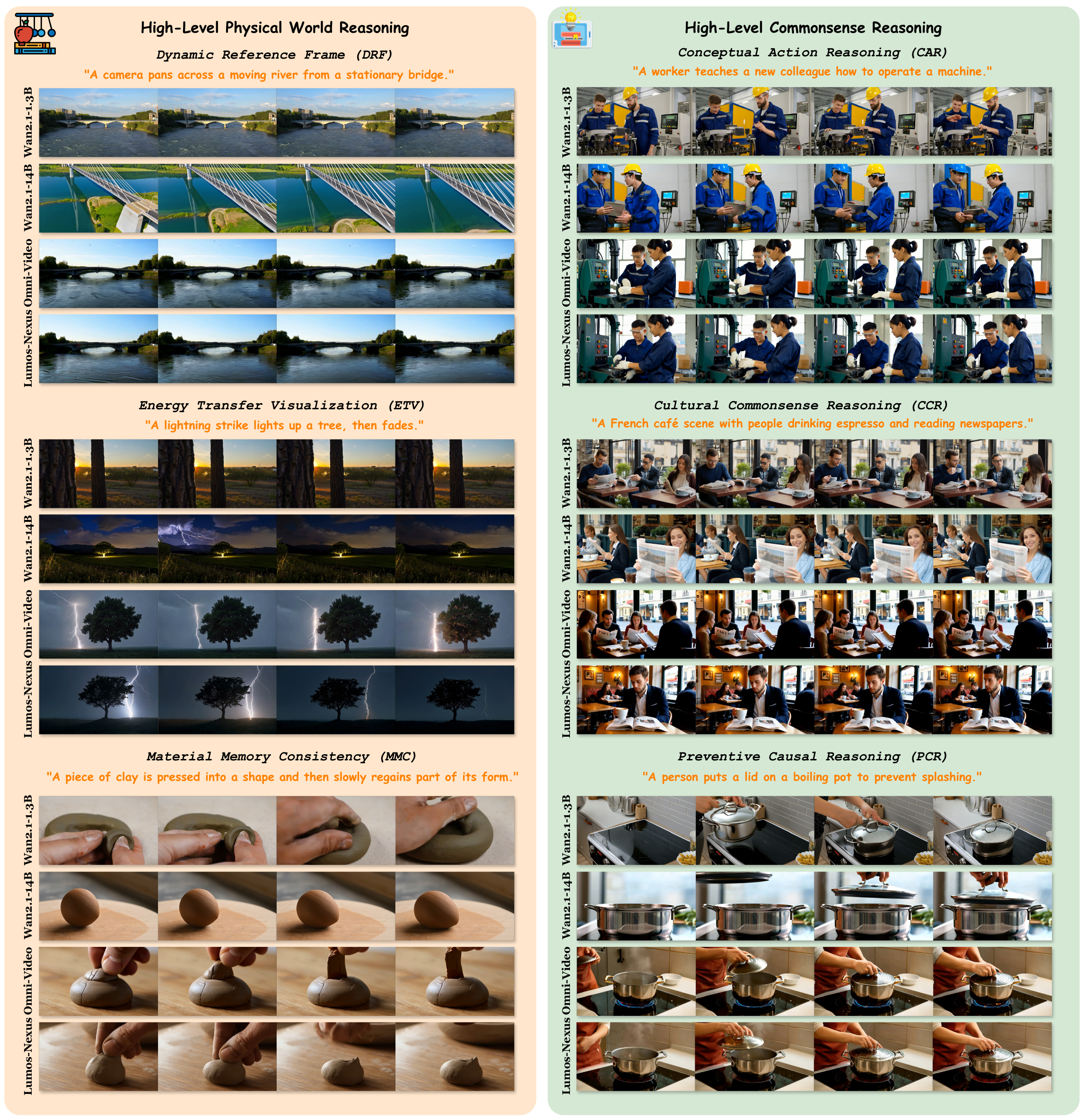}
  \caption{VR-Bench T2V qualitative comparison across two high-level reasoning categories: High-Level Physical World Reasoning, including Dynamic Reference Frame (DRF), Energy Transfer Visualization (ETV), and Material Memory Consistency (MMC), and High-Level Commonsense Reasoning, including Conceptual Action Reasoning (CAR), Cultural Commonsense Reasoning (CCR), and Preventive Causal Reasoning (PCR).}
  \label{fig:vis_app1}
\end{figure*}

\begin{figure*}[th!]
  \centering
  \includegraphics[width=\linewidth]{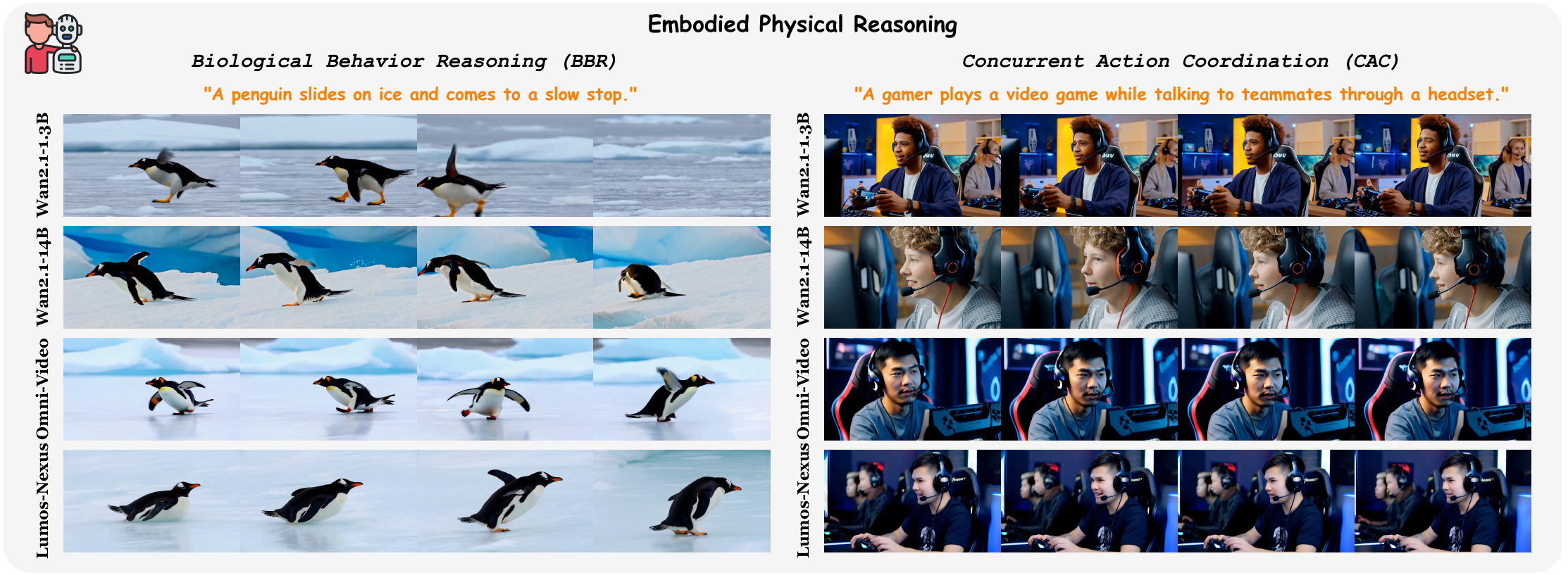}
  \caption{VR-Bench T2V qualitative comparison under the Embodied Physical Reasoning dimension, covering Biological Behavior Reasoning (BBR) and Concurrent Action Coordination (CAC ).}
  \label{fig:vis_app2}
\end{figure*}

% \subsection{Visualization Results}

% =========================================================
\subsection{Summary}
% =========================================================
Fig.~\ref{fig:vis_app1} and Fig.~\ref{fig:vis_app2} present quantitative comparison visualizations of different methods across all eight dimensions. In summary, VR-Bench provides a structured and interpretable framework for diagnosing whether a video generation model \emph{understands} the world it depicts.  
By jointly evaluating physical, causal, biological, and cultural reasoning, VR-Bench offers a comprehensive benchmark for assessing next-generation reasoning-driven video generation systems.

\section{Quantitative T2I Results}
\label{t2i-result}
As shown in Tab.~\ref{tab:t2i_geneval}, \ourbridge achieves the best overall performance (0.79) among unified models and outperforms all prior approaches across most evaluation dimensions in GenEval~\cite{ghosh2023geneval}. Compared with the baseline Omni-Video~\cite{tan2025omni} (0.75), \ourbridge delivers consistent gains in Two Objects, Colors, Position, and especially Attribute Binding (0.67 vs. 0.56), demonstrating stronger compositional understanding and more reliable instruction adherence. While generative models such as FLUX~\cite{flux2024} and SD3~\cite{2024SD3}, as well as unified models like MetaQuery~\cite{pan2025transfer}, demonstrate competitive performance, \ourbridge achieves overall stronger results, indicating that our training-efficient design—leveraging reasoning-guided semantics and UPFB-based refinement—achieves high-fidelity T2I generation while maintaining robust semantic controllability.

\begin{table*}[t]
    \centering
    \definecolor{lightblue}{RGB}{240,248,255}
    \newcommand{\colorrow}[1]{\rowcolor{lightgray} #1}
    \caption{
    {Performance comparison on GenEval.} The \textbf{boldfacen} and \underline{underline} font indicate the highest and the second highest results.
    % In the ``\#Params" column, we present the parameter counts of the visual generative model and the significant \textcolor{gray}{language encoders} for a fair comparison with unified models following~\cite{zhou2024transfusion}.
    % Tables below follow this paradigm.
    }
    % \vspace{-.3cm}
    \label{tab:t2i_geneval}
    \resizebox{0.8\linewidth}{!}{
        \begin{tabular}{l|c|cccccc}
            \toprule
            \textbf{Model} & \textbf{Overall$\uparrow$} & \textbf{Single Obj.} & \textbf{Two Obj.} & \textbf{Counting} & \textbf{Colors} & \textbf{Position} & \textbf{Attr. Bind} \\
            % & \textbf{Ext. Encoder}
            \midrule
            % \midrule
            \multicolumn{8}{l}{\textbf{Generation models}} \\ % Span 10 columns
            \midrule
            SD v1.5~\cite{rombach2022stable_diffusion} & 0.43  & 0.97 & 0.38 & 0.35 & 0.76 & 0.04 & 0.06 \\
            SD v2.1~\cite{rombach2022stable_diffusion} & 0.50  &  \underline{0.98} & 0.51 & 0.44 & 0.85 & 0.07 & 0.17 \\
            % 1.3B
            SD-XL~\cite{2023SDXL} & 0.55 & \underline{0.98} & 0.74 & 0.39 & 0.85 & 0.15 & 0.23 \\
            % 3.4B
            % 2.6B \textcolor{gray}{+ 0.4B}
            SD 3~\cite{2024SD3}   & 0.68 & \underline{0.98} &  0.84 & 0.66 & 0.74 & 0.40 & 0.43 \\
            % 12.7B (from Mint) 
            DALL-E 2~\cite{2022DALLE2}  & 0.52 & 0.94 & 0.66 & 0.49 & 0.77 & 0.10 & 0.19 \\
            % 6.5B janusflow, infinity
            % 4.2B \textcolor{gray}{+ 1.0B}
            % DALL-E 3~\cite{2023dalle3} & -- & -- & 0.67 & 0.96 & 0.87 & 0.47 & 0.83 & 0.43 & 0.45 \\
            FLUX~\cite{flux2024} & 0.67 & \textbf{0.99} & 0.85 & 0.75 & 0.77 & 0.22 & 0.42 \\
            Lumos-1~\cite{yuan2025lumos}  & 0.66 & {0.95} & 0.81 & 0.46 & 0.81 & 0.48 & 0.48 \\
            % Wan2.1-1.3B~\cite{wang2025wan} & 0.71 & {\underline{0.98}} & 0.82 & 0.54 & 0.83 & 0.55 & 0.55 \\
            % Wan2.1-14B~\cite{wang2025wan} & 0.77 & {0.99} & 0.89 & 0.76 & 0.86 & 0.55 & 0.60 \\
            % FLUX~\cite{flux2024} & 12B \textcolor{gray}{+ 2.5B} & -- & 0.665 & \underline{0.98}8 & 0.849 & 0.747 & 0.766 & 0.218 & 0.423 \\
            % IF-XL~\cite{2023IF} & 10.1B &  & 0.61 & 0.97 & 0.74 & 0.66 & 0.81 & 0.13 & 0.35 \\
            \midrule
            \multicolumn{8}{l}{\textbf{Unified models}} \\ % Span 10 columns
            \midrule
            LlamaGen~\cite{sun2024LlamaGen} & 0.32 & 0.71 & 0.34 & 0.21 & 0.58 & 0.07 & 0.04 \\
            Show-o~\cite{xie2024show-o}& 0.53 & 0.95 & 0.52 & 0.49 & 0.82 & 0.11 & 0.28 \\
            Chameleon~\cite{team2024Chameleon}  & 0.39 & -- & -- & -- & -- & -- & -- \\
            Transfusion~\cite{zhou2024transfusion} & 0.63 & -- & -- & -- & -- & -- & -- \\
            EMU3~\cite{zhou2024transfusion} & 0.66 & \textbf{0.99} & 0.81 & 0.42 & 0.80 & 0.49 & 0.45 \\
            LWM~\cite{liu2024LWM} & 0.47 & 0.93 & 0.41 & 0.46 & 0.79 & 0.09 & 0.15 \\
            Janus~\cite{wu2025janus}  &  0.61 & 0.97 & 0.68 & 0.30 & 0.84 & 0.46 & 0.42 \\
            JanusFlow~\cite{ma2025janusflow} & 0.63 & 0.97 & 0.59 & 0.45 & 0.83 & \textbf{0.53} & 0.42 \\
            SEED-X~\cite{ge2024seed-x}  & 0.49 & 0.97 & 0.58 & 0.26 & 0.80 & 0.19 & 0.14 \\
            MetaQuery-L~\cite{pan2025transfer} & \underline{0.78} & - & - & - & - & - & - \\
            Omni-Video~\cite{tan2025omni} & 0.75 & \textbf{0.99} & \underline{0.89} & \textbf{0.84} & \underline{0.87} & 0.35 & \underline{0.56} \\
            \ours  & \textbf{0.79} & \textbf{0.99} & \textbf{0.90} & \underline{0.82} & \textbf{0.88} & \underline{0.50} & \textbf{0.67} \\
            % \colorrow{Janus~\cite{2024Janus} & 1.3B &  & 0.61 & 0.97 & 0.68 & 0.30 & 0.84 & 0.46 & 0.42 \\}
            % \colorrow{JanusFlow~\cite{ma2024janusflow} & 1.3B &  & 0.63 & 0.97 & 0.59 & 0.45 & 0.83 & 0.53 & 0.42 \\}
            % \colorrow{\textbf{MINT (Ours)} & 1.3B &  & \textbf{0.73} & \textbf{\underline{0.98}} & \textbf{0.82} & \textbf{0.66} & 0.79 & \textbf{0.55} & \textbf{0.56} \\}
          
            % \method (Qwen32B caption, cfg = 16)
            \bottomrule
        \end{tabular}}
% \vspace{-6mm}
\end{table*}

% \begin{table}[t!]
% \caption{Performance comparison with and without RMS alignment and energy re-balancing in UPFB.}
%     \vspace{-5pt}

% \centering
% \resizebox{0.6\linewidth}{!}{
% \begin{tabular}{lccc}
% \toprule
% \textbf{Method} & \multicolumn{3}{c}{\textbf{VBench}} \\
% \cmidrule(lr){2-4}
%  & \textbf{Total Score} & \textbf{Quality Score} & \textbf{Semantic Score} \\
% \midrule
% w/o RMS  & 84.07 & 84.98  & {80.43}  \\
% w/ RMS  & \textbf{84.12 }& \textbf{85.03}  & \textbf{{80.52}}  \\
% \bottomrule
% \end{tabular}}
% \label{tab:rms}
%     \vspace{-5pt}

% \end{table}

% \begin{figure*}[ht!]
%   \centering
%   \includegraphics[width=\linewidth]{figs/abl_gammat.pdf}
%   \caption{Visualized qualitative comparison under varying $\gamma_w$ on $w_t$.}
%   \label{fig:abl_gmma_t}
% \end{figure*}

\section{More Ablation Discussion}
\label{app_ablation}

\subsection{Discussion on Bridging Large and Small Generators in the Video Unified Model}
\label{bridge_models}

\begin{figure}[ht!]
  \centering
  \includegraphics[width=0.9\linewidth]{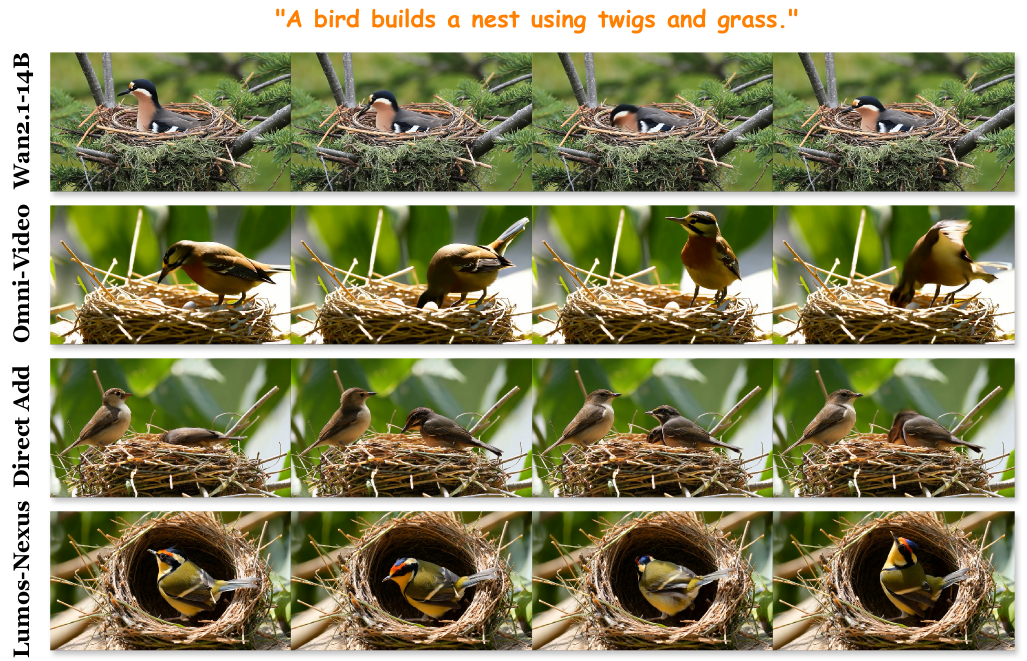}
  \vspace{-10pt}
  \caption{Qualitative visualization comparing different bridging strategies between small and large generators.}
  \label{fig:abl_bridge}
\end{figure}
Fig.~\ref{fig:abl_bridge} presents a qualitative comparison of different strategies for combining the small generator (Omni-Video~\cite{tan2025omni}) and the large generator (Wan-2.1-14B~\cite{wang2025wan}). The Direct Add baseline simply averages the noise predictions of the two generators at every sampling step. Notably, Direct Add frequently produces duplicated or structurally inconsistent subjects—for example, generating two birds (row 3) or even a single bird with two heads (row 3 col 3), despite the prompt specifying only one bird. These failures illustrate that uncontrolled blending disrupts both semantic grounding and spatial coherence. In contrast, our Lu mo s performs temporally scheduled and frequency-aware fusion, which respects each generator’s strengths.
This qualitative advantage is further supported by quantitative results: as shown in Tab.~\ref{tab:abl_quan_bridge}, Lumos-Nexus outperforms Direct Add by +0.84 on VBench (84.12 \textit{vs.} 83.28) and by +1.95 on VR-Bench (79.28 \textit{vs.} 77.33). The consistent gains across both benchmarks validate the effectiveness of our frequency-domain fusion and velocity decomposition. As a result, Lumos-Nexus produces cleaner structures, maintains single-subject consistency, and achieves significantly improved realism, demonstrating that our UPFB design provides a far more stable and semantically aligned bridging mechanism.
% This results in cleaner structures, single-subject consistency, and significantly improved realism, demonstrating that our UPFB design provides a far more stable and semantically aligned bridging mechanism.

\begin{table*}[t!]
\caption{Quantitative performance comparing different bridging strategies between small and large generators.}
    \vspace{-5pt}

\centering
\resizebox{\linewidth}{!}{
\begin{tabular}{lcccccccc}
\toprule
\textbf{Method} & \multicolumn{3}{c}{\textbf{VBench}} & \multicolumn{4}{c}{\textbf{VR-Bench}} \\
\cmidrule(lr){2-4} \cmidrule(lr){5-8}
 & \textbf{Total Score} & \textbf{Quality Score} & \textbf{Semantic Score} 
 & \textbf{Total Score} & \textbf{HL-Phys.} & \textbf{HL-Comm.} & \textbf{Emb-Phys.} \\
\midrule
Direct Add & 83.28  &  83.71 & 77.58 & 77.33 & {77.11}  & 75.89  & 79.82  \\
Lumos-Nexus & \textbf{84.12} & \textbf{85.03} & \textbf{80.52} & \textbf{79.28} & \textbf{79.49}  & \textbf{77.57} & \textbf{81.54} \\
\bottomrule
\end{tabular}}
\label{tab:abl_quan_bridge}
\end{table*}

\begin{table*}[t]
\centering
\caption{
Performance comparison on VR-Bench, evaluated by GPT 5.2.
The eight metrics are grouped into three reasoning categories:
High-Level Physical World Reasoning (HL-Phys.),
High-Level Commonsense Reasoning (HL-Comm.),
and Embodied Physical Reasoning (Emb.-Phys.).
\textbf{Bold} and \underline{underline} indicate the best and the second-best results, respectively.
}
\label{tab:reasoning_results_gpt}
\resizebox{\linewidth}{!}{
\begin{tabular}{l|cccc|cccccccc}
\toprule
\textbf{Model} 
& \textbf{Total$\uparrow$} 
& \textbf{HL-Phys.} 
& \textbf{HL-Comm.} 
& \textbf{Emb.-Phys.} 
& \textbf{DRF} 
& \textbf{ETV} 
& \textbf{MMC} 
& \textbf{CAR} 
& \textbf{CCR} 
& \textbf{PCR} 
& \textbf{BBR} 
& \textbf{CAC} \\
\midrule
CogVideoX1.5-5B~\cite{yang2024cogvideox} 
& 59.36 & 62.54 & 57.10 & 57.96 
& 76.67 & \textbf{57.20} & 53.75 & 57.76 & 70.74 & 42.80 & \underline{70.37} & 45.56 \\
HunyuanVideo~\cite{kong2024hunyuanvideo} 
& 66.15 & \underline{66.12} & 62.72 & \textbf{71.36} 
& \underline{76.91} & 55.33 & 66.11 & 64.73 & 72.22 & 51.20 & \textbf{72.10} & \underline{70.62} \\
Wan2.1-1.3B~\cite{wang2025wan} 
& 63.26 & 62.06 & 61.19 & 68.15 
& 72.72 & 53.47 & 60.00 & 66.34 & 67.90 & 49.33 & 65.68 & \underline{70.62} \\
Wan2.1-14B~\cite{wang2025wan} 
& \underline{67.22} & 65.16 & \underline{68.00} & 69.14 
& 74.94 & 53.60 & \underline{66.94} & \textbf{70.49} & \underline{77.65} & \underline{55.87} & 69.26 & 69.01 \\
Omni-Video~\cite{tan2025omni} 
& 62.04 & 61.42 & 61.64 & 63.58 
& 76.30 & 48.67 & 59.31 & 62.60 & 74.32 & 48.00 & 63.95 & 63.21 \\
\midrule
\ours 
& \textbf{68.51} 
& \textbf{67.36} 
& \textbf{68.63} 
& \underline{70.08} 
& \textbf{77.78} 
& \underline{56.93} 
& \textbf{67.36} 
& \underline{70.04} 
& \textbf{78.64} 
& \textbf{57.20} 
& 67.44 
& \textbf{72.72} \\
\bottomrule
\end{tabular}}
\end{table*}

\begin{table*}[t!]
\centering
\caption{\footnotesize{Comparison of inference/training cost and performance. Omni-Video* denotes the variant that replaces the original 1.3B generator with Wan2.1-14B and is fine-tuned via LoRA under the same framework.}}
\vspace{-10pt}
\setlength{\tabcolsep}{2pt}
\resizebox{0.9\linewidth}{!}{
\begin{tabular}{l c c c c c c c}
\hline
\multirow{2}{*}{Methods} &
\multicolumn{3}{c}{\textbf{Inference Cost (1 H20)}} &
\multicolumn{2}{c}{\textbf{Training Cost (8 H20s)}} &
\multirow{2}{*}{\textbf{VBench}} &
\multirow{2}{*}{\textbf{VR-Bench}} \\
\cmidrule(lr){2-4} \cmidrule(lr){5-6}
& Latency & FLOPs & GPU Usage & Latency & GPU Usage & & \\
\midrule
Wan2.1-14B~\cite{wang2025wan}    & 35.1 s/step & 3.462 P/step & 45.7 GB & 9.72 s/it & 94.7 GB & 83.69 & 78.23 \\
Omni-Video~\cite{tan2025omni}  & 7.8 s/step  & 1.014 P/step & 39.6 GB & 2.25 s/it & 26.3 GB & 83.82 & 72.78 \\
\textbf{Lumos-Nexus}   & 40.5 s/step & 4.568 P/step & 82.2 GB & 2.25 s/it & 26.3 GB & \textbf{84.12} & \textbf{79.28} \\ \hline
Omni-Video*  & 36.1 s/step & 3.513 P/step & 60.3 GB & 7.21 s/it & 72.8 GB & 81.73 & 70.63 \\
\hline
\end{tabular}}
\vspace{-10pt}

\label{tab:cost_perf}
\end{table*}

\subsection{VR-Bench Cross-Evaluator Robustness}\label{cross_eval}
To verify that VR-Bench results are robust to the choice of evaluator, we additionally conduct VR-Bench evaluation using a different judge model. Specifically, we replace Qwen3-VL-30B-A3B-Instruct~\cite{yang2025qwen3} with GPT-5.2 as the evaluator on VR-Bench. As shown in Tab.~\ref{tab:reasoning_results_gpt}, we re-evaluate the main baseline methods reported in Tab. 2 of the main text. The overall performance ranking is largely consistent between evaluations using GPT-5.2 and those using Qwen3-VL. Notably, Lumos-Nexus remains the top-performing method among the major open-source baselines across evaluators, supporting the stability of our conclusions under different evaluator choices.

\subsection{Discussion of Model Efficiency}
\label{diss_effi}
We report detailed inference and training cost comparisons across different models in Tab.~\ref{tab:cost_perf}. While Lumos-Nexus runs both the small and large generators at inference, its overall inference cost is only about 1.2× that of the large model alone (Wan2.1-14B~\cite{wang2025wan}) in terms of latency (40.5 \textit{vs.} 35.1 s/step), with a moderate increase in FLOPs (4.568P \textit{vs.} 3.462P per step). Despite this overhead, Lumos-Nexus achieves the best overall performance, outperforming Wan2.1-14B by +0.43 on VBench (84.12 \textit{vs.} 83.69) and +1.05 on VR-Bench (79.28 \textit{vs.} 78.23), demonstrating that the additional inference cost translates directly into measurable quality gains.
More importantly, we argue that training cost is the dominant bottleneck in unified video generation models. Training the large generator is substantially more expensive than the small one—requiring 9.72 s/it and 94.7GB GPU memory, compared to 2.25 s/it and 26.3GB for Omni-Video. In contrast, Lumos-Nexus fully aligns with the training cost of the small generator (2.25 s/it, 26.3GB), as it avoids any large-model training and only optimizes lightweight bridging modules. This means that while we incur a modest inference overhead, we completely eliminate the prohibitive cost of large-scale video model training.
We believe this trade-off—slightly higher inference cost for large and consistent performance gains, while retaining small-model-level training cost—is both practical and highly scalable in real-world deployments.
 
 % We report detailed inference and training cost comparisons across different models in Tab.~\ref{tab:cost_perf}. While Lumos-Nexus runs both the small and large generators at inference, its overall inference cost is only about 1.2× that of the large model alone (Wan2.1-14B~\cite{wang2025wan}), yet it brings consistent gains on VBench and VR-Bench. More importantly, we argue that training cost is the dominant bottleneck in video unified models. Training the large generator is nearly 4× more expensive than the small one, whereas Lumos-Nexus fully aligns with the training cost of the small generator, avoiding any large-model training. We believe this trade-off is practical and scalable in real-world settings.

\subsection{Disentangling Model Capacity from UPFB Gains}
\label{UPFB_gains}
To examine whether the performance gains of Lumos-Nexus arise from increased generator capacity or from the proposed UPFB mechanism, we conduct a controlled comparison by upgrading the generator backbone in Omni-Video to the same 14B scale. Specifically, we replace the original Wan2.1-1.3B generator with Wan2.1-14B and fine-tune it via LoRA under the same training framework.
This variant, denoted as Omni-Video* in Tab.~\ref{tab:cost_perf}, is trained on 160K T2V samples using 8 H20 GPUs for nearly four days. As shown in the table, this upgrade significantly increases both training and inference cost: the training latency rises from 2.25 s/it to 7.21 s/it, and GPU memory usage increases from 26.3GB to 72.8GB. Inference cost also grows notably (36.1 s/step and 3.513P FLOPs), approaching that of the standalone 14B model.
However, despite this substantial increase in computational cost, Omni-Video* still underperforms Lumos-Nexus on both VBench (81.73 \textit{vs.} 84.12) and VR-Bench (70.63 \textit{vs.} 79.28). This clearly indicates that simply scaling up the generator backbone does not translate into better performance within the Omni-Video framework. We attribute this gap to architectural incompatibility rather than insufficient capacity: Wan2.1-14B cannot natively accept VLM tokens, making effective cross-modal alignment difficult under limited fine-tuning. Lightweight LoRA adaptation is insufficient to resolve this structural mismatch.
In contrast, Lumos-Nexus integrates Wan2.1-14B through the proposed UPFB mechanism without any additional large-model training, while maintaining the same training cost as the small generator (2.25 s/it, 26.3GB). The superior performance achieved under strictly controlled capacity and cost conditions demonstrates that the observed gains stem from the UPFB design itself, rather than from increased model scale.

\subsection{More Discussions of Hyperparameters}
\label{more_hyper}
We conduct a sensitivity study on two representative transition hyperparameters, $\gamma_w$ and $(\sigma_{\min}, \sigma_{\max})$, for Lumos-Nexus* with Wan2.2-T2V-A14B. As shown in Tab.~\ref{tab:wan22_hyperparams}, the overall trend is consistent with the Wan2.1 setting. For $\gamma_w$, the model achieves the best VR-Bench score of 81.90 when $\gamma_w=0.3$, while both smaller and larger values lead to performance drops. In particular, increasing $\gamma_w$ to 0.4 or 0.5 noticeably degrades the score, suggesting that overly strong transition weighting may disturb the generation quality. For $(\sigma_{\min}, \sigma_{\max})$, the setting $(0.35, 0.70)$ also obtains the best score of 81.90, outperforming both the smaller range $(0.05, 0.10)$ and the larger range $(1.00, 2.00)$. These results indicate that a moderate transition strength and noise range are important for stable performance, and further support the robustness of our hyperparameter choices across different Wan backbones.
\begin{table}[t!]
\centering
\small
\caption{Hyperparameter Sensitivity of Lumos-Nexus* with Wan2.2-T2V-A14B.}
\label{tab:wan22_hyperparams}
\resizebox{0.85\linewidth}{!}{
\begin{minipage}{0.95\linewidth}
\centering
\begin{tabular*}{\linewidth}{@{\extracolsep{\fill}} c c c c c @{}}
\hline
$\gamma_w$ & $0.2$ & $0.3$ & $0.4$ & $0.5$ \\
\hline
VR-Bench & 81.65 & \textbf{81.90} & 79.95 & 78.34 \\
\hline
\end{tabular*}

\begin{tabular*}{\linewidth}{@{\extracolsep{\fill}} c c c c @{}}
$(\sigma_{\min},\sigma_{\max})$ 
& $(0.05,0.10)$ 
& $(0.35,0.70)$ 
& $(1.00,2.00)$ \\
\hline
VR-Bench 
& 81.21
& \textbf{81.90}
& 80.95 \\
\hline
\end{tabular*}

\end{minipage}}
% \caption{VR-Bench scores evaluated by different evaluators.}
\label{tab:vrbench_evaluator}
\end{table}

\subsection{Human Evaluation on VR-Bench}
\label{human_eval}
We conduct human evaluation on 20 randomly sampled VR-Bench videos with 13 annotators, and report Qwen3-VL* scores on the same subset for reference. As shown in Tab.~\ref{tab:human_eval}, Lumos-Nexus consistently ranks first under both Qwen3-VL* and human evaluation, achieving 86.19 and 69.33, respectively. To quantify the ranking consistency, we further compute Kendall’s $\tau$~\cite{kendall1938new}, a standard rank correlation coefficient that measures the agreement between two orderings. The resulting $\tau=0.73$ indicates strong positive agreement between Qwen3-VL*-based rankings and human preferences, supporting the reliability of Qwen3-VL-based VR-Bench evaluation.
% We conduct human evaluation on 20 randomly sampled VR-Bench videos with 13 annotators, and report Qwen3-VL* on the same subset for reference. Across Qwen3-VL* and human evaluation, Lumos-Nexus consistently ranks first.
% Kendall's $\tau$ shows strong ranking agreement, supporting the reliability of Qwen3-VL-based VR-Bench results.

\begin{figure}[t!]
  \centering
  \includegraphics[width=0.9\linewidth]{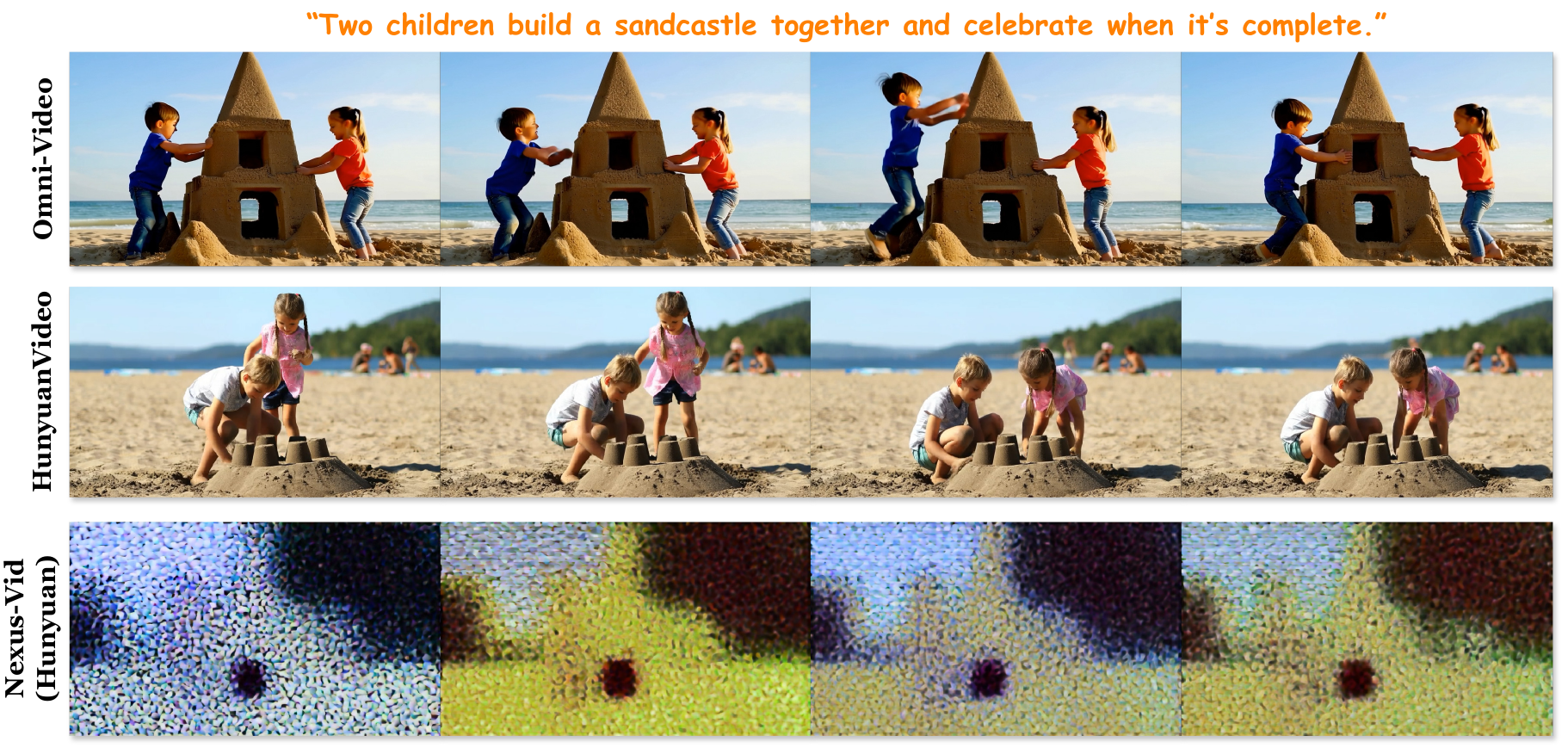}
  \vspace{-10pt}
  \caption{Qualitative visualization of replacing the large generator in Lumos-Nexus with a heterogeneous video generation model.}
  \label{fig:yigou}
\end{figure}

\begin{table*}[t!]
\centering
\caption{Human Evaluation on VR-Bench. K-$\tau$ denotes Kendall’s rank correlation coefficient between the two rankings.}
\small
\setlength{\tabcolsep}{3pt}
\resizebox{\linewidth}{!}{
\begin{tabular}{l c c c c c c c}
\hline
& \textbf{CogVideoX}~\cite{yang2024cogvideox} & \textbf{HunyuanVideo}~\cite{kong2024hunyuanvideo} & \textbf{Wan2.1-1.3B}~\cite{wang2025wan} & \textbf{Wan2.1-14B}~\cite{wang2025wan} & \textbf{Omni-Video}~\cite{tan2025omni} & \textbf{Lumos-Nexus} & {K-$\tau$} \\
\hline
Qwen3-VL* & 57.95 & 83.42 & 79.45 & 83.28 & 82.63 & \textbf{86.19} & \multirow{2}{*}{0.73} \\
Human   & 50.34 & 66.48 & 66.32 & 67.94 & 67.85 & \textbf{69.33} &  \\
\hline
\end{tabular}}
\vspace{-13pt}
% \caption{VR-Bench scores evaluated by different evaluators.}
\label{tab:human_eval}
\end{table*}

\begin{table}[t]
\centering
\caption{Maximum Mean Discrepancy (MMD) between the latents of Wan2.1-T2V-1.3B~\cite{wang2025wan} and different video . Lower values indicate better alignment.}
\small
\resizebox{\linewidth}{!}{
\begin{tabular}{l c c c c c}
\hline
 & \textbf{Wan2.1-T2V-14B}~\cite{wang2025wan} & \textbf{Wan2.2-T2V-A14B}~\cite{wang2025wan}& \textbf{Wan2.2-TI2V-5B}~\cite{wang2025wan} & \textbf{HunyuanVideo}~\cite{kong2024hunyuanvideo}  \\
\hline
Wan2.1-T2V-1.3B~\cite{wang2025wan} & 0.523 & 0.531 & 2.977 & 1.031 \\
\hline
\end{tabular}}
\label{tab:mmd}
\end{table}

\subsection{Discussion of the Homogeneous Latent Space Assumption}
\label{assump_latent_space}
Our framework is built on the assumption of a homogeneous latent space, meaning that the generators involved share a compatible VAE latent space such that their noise or velocity predictions remain distributionally aligned. This property is important for maintaining consistent generation behavior across models. When this assumption does not hold, the mismatch in latent representations can significantly impair performance. For example, as shown in Fig.~\ref{fig:yigou} (row 3), replacing the large generator in Lumos-Nexus with a heterogeneous model such as HunyuanVideo~\cite{kong2024hunyuanvideo} results in severely degraded and blurry outputs, which can be attributed to misaligned noise prediction and incompatible latent spaces.
This assumption is reasonable in many practical scenarios. Modern generative model families often provide multiple scale variants that are developed within a unified latent-space design and share the same VAE. Examples include model variants such as Wan2.1-T2V~\cite{wang2025wan} and Wan2.2-T2V-A14B~\cite{wang2025wan}, which makes them naturally compatible within our framework. Even when heterogeneous models are considered, they can still be incorporated through training-stage alignment, where the small generator is adapted to match the latent space of the large generator. Compared with retraining the large generator, this strategy is substantially more efficient.
To quantify the degree of latent-space compatibility, we measure the discrepancy between model latent distributions using Maximum Mean Discrepancy (MMD), where lower values indicate better alignment. As shown in Tab.~\ref{tab:mmd}, the MMD between Wan2.1-T2V-1.3B and Wan2.1-T2V-14B, Wan2.2-T2V-A14B, Wan2.2-5B, and HunyuanVideo is 0.523, 0.531, 2.977, and 1.031, respectively. Notably, the latter two models do not share the same VAE with Wan2.1-1.3B, and both also exhibit substantially degraded performance when used as large generators. These results suggest that latent-space homogeneity is a key factor underlying the effectiveness of the proposed framework.
% To examine whether the performance gains of Lumos-Nexus can be attributed to increased generator capacity or to the proposed UPFB mechanism, we conduct a controlled comparison by upgrading the generator backbone in Omni-Video to the same 14B scale.
% We replace the original Wan2.1-1.3B generator in Omni-Video with Wan2.1-14B and fine-tune it using LoRA under the same framework. This variant, denoted as Omni-Video* in Tab.~\ref{tab:cost_perf}, is trained on 160K T2V samples using 8 H20 GPUs for nearly four days. Despite substantially higher training and inference costs, Omni-Video* still underperforms Lumos-Nexus, mainly because Wan2.1-14B cannot natively accept VLM tokens, making effective alignment difficult with limited training. In contrast, Lumos-Nexus integrates Wan2.1-14B through the proposed UPFB without additional training, demonstrating that the observed gains stem from UPFB rather than increased model capacity.

\subsection{Discussion of Lumos-Nexus's Generality}
\label{generality}
As discussed in Appendix~\ref{assump_latent_space}, Lumos-Nexus requires a homogeneous latent space, yet this requirement is not specific to the Wan family. To examine its generality, we further test Lumos-Nexus on CogVideoX-2B and CogVideoX-5B, which also satisfy this condition. Due to time constraints, we do not retrain CogVideoX-2B with an attached VLM under the full Lumos-Nexus training pipeline. Instead, we directly bridge the two models at inference time using the same prompt.
As shown in Tab.~\ref{tab:cogvideox_generality}, the resulting CogVideoX* can generate valid videos and achieves a VR-Bench score of 65.79, which falls between CogVideoX-2B and CogVideoX-5B. This expected trend suggests that applying VLM-aligned CogVideoX-2B training within Lumos-Nexus could further improve CogVideoX*, potentially surpassing CogVideoX-5B.
% \subsection{Discussion of Lumos-Nexus's Generality}
% \label{generality}
%  As discussed in Appendix~\ref{assump_latent_space}, Lumos-Nexus requires a homogeneous latent space, but this assumption is not limited to the Wan family. We verify this on CogVideoX-2B and CogVideoX-5B, which meet this condition. Due to time constraints, we do not retrain CogVideoX-2B after attaching a VLM within Lumos-Nexus. Instead, we directly bridge the two models at inference using the same prompt.
%  As shown in the table below, CogVideoX* generates valid videos and achieves a VR-Bench score between CogVideoX-2B and CogVideoX-5B, as expected. 
% This trend suggests that VLM-aligned CogVideoX-2B training within Lumos-Nexus could further improve CogVideoX* beyond CogVideoX-5B.

\section{Q\&A Evaluation Examples from VR-Bench}
\label{QAs_VR-Bench}
We provide detailed question–answer evaluations for each reasoning dimension in VR-Bench, as illustrated below. For every dimension, we select a representative case and perform the full eight-question diagnostic assessment on the video generated by our \ourbridge. Each selected case aligns exactly with the examples presented in Fig.~\ref{fig:vis_app1} and Fig.~\ref{fig:vis_app2}, allowing for transparent, fine-grained inspection of our model’s capability in physical, commonsense, and embodied reasoning. These expanded Q\&A results offer clear evidence of the reasoning consistency achieved by our approach across all dimensions of VR-Bench.

\begin{table}[t]
\centering
\small
\caption{Generality Evaluation of Lumos-Nexus on CogVideoX.}
\label{tab:cogvideox_generality}
\setlength{\tabcolsep}{5pt}
\resizebox{0.625\linewidth}{!}{
\textbf{}\begin{tabular}{l c c c}
\hline
 & \textbf{CogVideoX-2B~\cite{yang2024cogvideox}} & \textbf{CogVideoX-5B~\cite{yang2024cogvideox}} & \textbf{CogVideoX*}  \\
\hline
VR-Bench & 64.88 & \textbf{66.27} & 65.79 \\
\hline
\end{tabular}}
\vspace{-13pt}
% \caption{VR-Bench scores evaluated by different evaluators.}
\label{tab:vrbench_evaluator}
\end{table}

\section{Limitaion}
\label{limit_label}
While \ourbridge effectively enhances visual fidelity and further strengthens reasoning-driven generative capability, it still inherits several limitations. VR-Bench, while comprehensive, cannot fully cover the open-world diversity of real-world reasoning scenarios. Broader categories of reasoning, such as long-horizon causal chains, remain unexplored. We leave deeper exploration within video unified models, as well as broader coverage of reasoning dimensions in our benchmark, for future work.

% =========================
% Figure: DRF only
% =========================
\begin{tcolorbox}[enhanced,breakable,myboxstyle,title={High-Level Physical World Reasoning}]
\begin{tcolorbox}[enhanced,breakable,drfstyle,title={DRF Evaluation}]
{\small  % ← 整体字体变小
\setlength{\parskip}{2pt}

% -------- Scene Viewpoint Consistency --------
\noindent\textbf{\textcolor{purple}{Scene Viewpoint Consistency}}
\begin{itemize}[leftmargin=15pt,itemsep=4pt]

  \item \textcolor{blue}{Q1}:  
  Is the viewpoint clearly identifiable as being inside or associated with a moving system (stationary bridge)?  
  \\[-0.2em]
  \textbf{Ans:} Yes \quad
  {\color{gray}\textbf{Reason:} The camera's perspective remains fixed relative to the bridge, which is a stationary structure. The bridge serves as a central, unchanging reference point in the frame, indicating the viewpoint is associated with it.}

  \item \textcolor{blue}{Q2}:  
  Does the camera maintain a consistent perspective relative to the stationary bridge throughout the pan?
  \\[-0.2em]
  \textbf{Ans:} Yes \quad
  {\color{gray}\textbf{Reason:} The camera consistently maintains its perspective relative to the bridge. The bridge remains the central subject, with stable position and orientation throughout the sequence, confirming viewpoint consistency.}

\end{itemize}

% -------- Relative Motion Coherence --------
\noindent\textbf{\textcolor{purple}{Relative Motion Coherence}}
\begin{itemize}[leftmargin=15pt,itemsep=4pt]

  \item \textcolor{blue}{Q3}:  
  Are relative motion directions between observer and background correct (unidirectional river flow)?  
  \\[-0.2em]
  \textbf{Ans:} Yes \quad
  {\color{gray}\textbf{Reason:} The river flows from the background toward the foreground, aligning with the expected motion during a camera pan. The water movement is unidirectional and consistent with a left-to-right camera pan.}

  \item \textcolor{blue}{Q4}:  
  Is the motion speed and direction physically plausible (unidirectional river flow)?  
  \\[-0.2em]
  \textbf{Ans:} Yes \quad
  {\color{gray}\textbf{Reason:} The river flows smoothly and continuously in a single direction toward the camera. The speed appears stable and physically plausible, consistent with natural river flow.}

  \item \textcolor{blue}{Q5}:  
  Is parallax or motion depth correctly represented (unidirectional river flow)?  
  \\[-0.2em]
  \textbf{Ans:} Yes \quad
  {\color{gray}\textbf{Reason:} The foreground water moves faster than the background, producing a clear parallax effect. This correctly reflects depth-based motion with a consistent flow pattern across the river surface.}

\end{itemize}

% -------- Physical Temporal Realism --------
\noindent\textbf{\textcolor{purple}{Physical Temporal Realism}}
\begin{itemize}[leftmargin=15pt,itemsep=4pt]

  \item \textcolor{blue}{Q6}:  
  Does the river appear to flow smoothly in the direction consistent with the camera's pan movement throughout the video?
  \\[-0.2em]
  \textbf{Ans:} Yes \quad
  {\color{gray}\textbf{Reason:} The river flows smoothly from the background toward the foreground, consistent with the camera's left-to-right pan. The motion remains continuous and aligned with the pan direction.}

  \item \textcolor{blue}{Q7}:  
  Are motions temporally consistent and smooth without abrupt reversals (static camera realism)?  
  \\[-0.2em]
  \textbf{Ans:} Yes \quad
  {\color{gray}\textbf{Reason:} The camera movement is smooth and continuous, without abrupt changes in speed or direction. The motion appears stable and realistic, resembling a steady pan.}

  \item \textcolor{blue}{Q8}:  
  Do lighting and reflections stay coherent  with the direction of motion (static camera realism)?  
  \\[-0.2em]
  \textbf{Ans:} Yes \quad
  {\color{gray}\textbf{Reason:} The lighting and reflections on the water remain consistent with the direction of motion. Reflections of the sky and trees shift smoothly across the surface, aligning with the continuous pan.}

\end{itemize}

\vspace{1ex}
\textbf{{Weighted Average Score}}: 1.00
}
\end{tcolorbox}
\end{tcolorbox}

% =========================
% Figure: ETV only
% =========================
\begin{tcolorbox}[enhanced,breakable,myboxstyle,title={High-Level Physical World Reasoning}]
\begin{tcolorbox}[enhanced,breakable,drfstyle,title={ETV Evaluation}]
{\small
\setlength{\parskip}{2pt}

% -------- Energy Continuity --------
\noindent\textbf{\textcolor{purple}{Energy Continuity}}
\begin{itemize}[leftmargin=15pt,itemsep=4pt]

  \item \textcolor{blue}{Q1}: Is the lightning discharge instantaneous and directional?  
  \\[-0.2em]
  \textbf{Ans:} Yes \quad
  {\color{gray}\textbf{Reason:} The lightning appears as a sudden, bright flash originating from the sky and striking the ground, which is characteristic of an instantaneous discharge. The path of the lightning is clearly directional, traveling from the upper atmosphere to the ground.}

  \item \textcolor{blue}{Q2}: Does the tree illumination occur immediately after strike?  
  \\[-0.2em]
  \textbf{Ans:} Yes \quad
  {\color{gray}\textbf{Reason:} The tree is dark in the initial frames. In the first frame where the lightning is visible, the tree is immediately illuminated by the flash, indicating that the illumination occurs concurrently with the strike.}

\end{itemize}

% -------- Directional Consistency --------
\noindent\textbf{\textcolor{purple}{Directional Consistency}}
\begin{itemize}[leftmargin=15pt,itemsep=4pt]

  \item \textcolor{blue}{Q3}: Is brightness proportional to the energy intensity?  
  \\[-0.2em]
  \textbf{Ans:} Yes \quad
  {\color{gray}\textbf{Reason:} The brightness of the lightning is extremely high, and the light it casts on the tree and the surrounding landscape is intense, which is proportional to the immense energy of a lightning strike.}

  \item \textcolor{blue}{Q4}: Does the light fade naturally over time?  
  \\[-0.2em]
  \textbf{Ans:} Yes \quad
  {\color{gray}\textbf{Reason:} The video shows a clear progression from a bright flash to a gradual dimming. The light from the lightning fades over several frames, which is a natural and realistic attenuation of light energy.}

  \item \textcolor{blue}{Q5}: Does the brightness of the lightning accurately reflect the energy transfer in the scene? 
  \\[-0.2em]
  \textbf{Ans:} Yes \quad
  {\color{gray}\textbf{Reason:} The lightning is depicted as the most intense light source in the scene, and its brightness is consistent with the immense energy transfer it represents. The light illuminates the entire landscape, accurately reflecting the scale of the energy event.}

\end{itemize}

% -------- Physical Plausibility --------
\noindent\textbf{\textcolor{purple}{Physical Plausibility}}
\begin{itemize}[leftmargin=15pt,itemsep=4pt]

  \item \textcolor{blue}{Q6}: Is energy flow path from sky to tree continuous and visible?  
  \\[-0.2em]
  \textbf{Ans:} Yes \quad
  {\color{gray}\textbf{Reason:} The lightning bolt is a continuous, visible path of energy that travels from the sky down to the ground, clearly showing the energy flow path. The light from the strike also illuminates the tree, indicating the transfer of energy to the object.}

  \item \textcolor{blue}{Q7}: Does the environment react correctly (e.g., lighting of surroundings)?  
  \\[-0.2em]
  \textbf{Ans:} Yes \quad
  {\color{gray}\textbf{Reason:} The flash of lightning illuminates the entire scene, including the tree, the ground, and the distant horizon. This correct reaction shows the light spreading out from the source, lighting up the environment.}

  \item \textcolor{blue}{Q8}: Is energy attenuation smooth without visual discontinuity?  
  \\[-0.2em]
  \textbf{Ans:} Yes \quad
  {\color{gray}\textbf{Reason:} The fading of the light is smooth and continuous across the frames. There are no abrupt jumps or visual discontinuities; the brightness decreases gradually, which is consistent with the natural dissipation of light energy.}

\end{itemize}

\vspace{1ex}
\textbf{{Weighted Average Score}}: 1.00
}
\end{tcolorbox}
\end{tcolorbox}

\begin{tcolorbox}[enhanced,breakable,myboxstyle,title={High-Level Physical World Reasoning}]
% =====================================================
% === MMC Evaluation Sub-box
% =====================================================
\begin{tcolorbox}[enhanced,breakable,drfstyle,title={MMC Evaluation}]
{\small

% -------- Deformation Realism --------
\noindent\textbf{\textcolor{purple}{Deformation Realism}}
\begin{itemize}[leftmargin=15pt,itemsep=4pt]

  \item \textcolor{blue}{Q1}: Is the clay visibly deformed by pressure?
  \\[-0.2em]
  \textbf{Ans:} Yes  \quad
  \textbf{\textcolor{gray}{Reason:}} \textcolor{gray}{The clay is clearly deformed by the pressure applied by the fingers, creating a distinct indentation and a new shape.}

  \item \textcolor{blue}{Q2}: Is the contact region physically plausible?
  \\[-0.2em]
  \textbf{Ans:} Yes  \quad
  \textbf{\textcolor{gray}{Reason:}} \textcolor{gray}{The contact region shows a natural indentation and slight smearing of the clay, which is physically plausible for a soft, malleable material under pressure.}

\end{itemize}

% -------- Recovery Dynamics --------
\noindent\textbf{\textcolor{purple}{Recovery Dynamics}}
\begin{itemize}[leftmargin=15pt,itemsep=4pt]

  \item \textcolor{blue}{Q3}: Does the clay show gradual partial recovery?
  \\[-0.2em]
  \textbf{Ans:} Yes \quad
  \textbf{\textcolor{gray}{Reason:}} \textcolor{gray}{After the hand is removed, the clay is observed to slowly regain some of its original form, particularly around the edges of the indentation, indicating partial recovery.}

  \item \textcolor{blue}{Q4}: Is recovery slower than elastic materials?
  \\[-0.2em]
  \textbf{Ans:} Yes  \quad
  \textbf{\textcolor{gray}{Reason:}} \textcolor{gray}{The recovery process is not instantaneous; it occurs over a few frames, which is consistent with the slow, viscous flow of clay, unlike the immediate rebound of an elastic material.}

  \item \textcolor{blue}{Q5}: Does residual deformation remain visible?
  \\[-0.2em]
  \textbf{Ans:} Yes \quad
  \textbf{\textcolor{gray}{Reason:}} \textcolor{gray}{A visible indentation remains in the clay after the hand is removed, demonstrating that the deformation is not fully elastic and that some permanent deformation has occurred.}

\end{itemize}

% -------- Material Trace Consistency --------
\noindent\textbf{\textcolor{purple}{Material Trace Consistency}}
\begin{itemize}[leftmargin=15pt,itemsep=4pt]

  \item \textcolor{blue}{Q6}: Does the recovered shape maintain the overall structural integrity of the pressed form?
  \\[-0.2em]
  \textbf{Ans:} Yes  \quad
  \textbf{\textcolor{gray}{Reason:}} \textcolor{gray}{Despite the indentation, the overall shape of the clay piece remains intact and recognizable, maintaining its structural integrity as a single, cohesive object.}

  \item \textcolor{blue}{Q7}: Is recovery visually consistent with clay’s viscosity?
  \\[-0.2em]
  \textbf{Ans:} Yes \quad
  \textbf{\textcolor{gray}{Reason:}} \textcolor{gray}{The slow, gradual nature of the recovery, where the clay slowly flows back, is consistent with the high viscosity and plasticity of clay.}

  \item \textcolor{blue}{Q8}: Does the motion suggest material memory retention?
  \\[-0.2em]
  \textbf{Ans:} Yes  \quad
  \textbf{\textcolor{gray}{Reason:}} \textcolor{gray}{The persistence of the indentation and the slow, incomplete recovery suggest that the material retains a 'memory' of the applied force, which is a key characteristic of plastic deformation.}

\end{itemize}
\vspace{1ex}
\textbf{{Weighted Average Score}}: 1.00
}
\end{tcolorbox}

\end{tcolorbox}

% =========================
% Figure: CCR only
% =========================
\begin{tcolorbox}[enhanced,breakable,myboxstyle,title={High-Level Commonsense Reasoning}]
\begin{tcolorbox}[enhanced,breakable,drfstyle,title={CCR Evaluation}]
{\small

% =====================================================
% Cultural Symbol Understanding
% =====================================================
\noindent\textbf{\textcolor{purple}{Cultural Symbol Understanding}}
\begin{itemize}[leftmargin=15pt,itemsep=4pt]

  \item \textcolor{blue}{Q1}:  
  Is the café outdoor with small round tables and chairs typical of France?  
  \\[-0.2em]
  \textbf{Ans:} No \quad
  {\color{gray}\textbf{Reason}: The café is indoors, not outdoors. The tables are rectangular, not small round tables typical of French sidewalk cafés.}

  \item \textcolor{blue}{Q2}:  
  Are people drinking espresso or similar small coffee cups, not mugs?  
  \\[-0.2em]
  \textbf{Ans:} Yes \quad
  {\color{gray}\textbf{Reason}: The man in the foreground is holding a small white cup, consistent with an espresso or a similar small coffee drink, which is typical for a French café.}

\end{itemize}

% =====================================================
% Contextual Alignment
% =====================================================
\noindent\textbf{\textcolor{purple}{Contextual Alignment}}
\begin{itemize}[leftmargin=15pt,itemsep=4pt]

  \item \textcolor{blue}{Q3}:  
  Is the body posture relaxed, suggesting conversation or solitude?  
  \\[-0.2em]
  \textbf{Ans:} Yes \quad
  {\color{gray}\textbf{Reason}: The individuals in the scene are sitting in a relaxed posture, focused on their own activities like reading and drinking, suggesting a calm, solitary, or quiet conversational atmosphere.}

  \item \textcolor{blue}{Q4}:  
  Are the people dressed in typical Parisian café attire, such as scarves or berets?  
  \\[-0.2em]
  \textbf{Ans:} No \quad
  {\color{gray}\textbf{Reason}: The people are dressed in business casual attire (suits, shirts). There are no visible scarves or berets, which are not typical everyday wear for most people in a modern Parisian café.}

  \item \textcolor{blue}{Q5}:  
  Are newspapers, croissants, or baguettes visible as cultural cues?  
  \\[-0.2em]
  \textbf{Ans:} Yes \quad
  {\color{gray}\textbf{Reason}: An open newspaper is clearly visible on the table in front of the main subject, serving as a cultural cue. While croissants or baguettes are not visible, the newspaper is a strong cultural indicator.}

\end{itemize}

% =====================================================
% Social Behavior Understanding
% =====================================================
\noindent\textbf{\textcolor{purple}{Social Behavior Understanding}}
\begin{itemize}[leftmargin=15pt,itemsep=4pt]

  \item \textcolor{blue}{Q6}:  
  Are the people in the scene appear to be engaged in typical café activities like drinking coffee and reading newspapers simultaneously?  
  \\[-0.2em]
  \textbf{Ans:} Yes \quad
  {\color{gray}\textbf{Reason}: The man in the foreground is simultaneously holding a coffee cup and reading a newspaper. In the background, another person is also reading a newspaper. This accurately depicts typical café activities.}

  \item \textcolor{blue}{Q7}:  
  Do the people in the scene appear to be enjoying their drinks and reading the newspapers in a relaxed manner consistent with a typical café setting?  
  \\[-0.2em]
  \textbf{Ans:} Yes \quad
  {\color{gray}\textbf{Reason}: The individuals are engaged in quiet, leisurely activities. The man's focused gaze on the newspaper and his relaxed hold on the cup suggest he is enjoying a moment of calm, consistent with a typical café setting.}

  \item \textcolor{blue}{Q8}:  
  Do the people in the scene appear to be interacting with their newspapers or drinks in a manner consistent with typical café behavior?  
  \\[-0.2em]
  \textbf{Ans:} Yes \quad
  {\color{gray}\textbf{Reason}: The man is holding the coffee cup and has his hand near the newspaper, indicating active engagement with both items. The other patrons are also actively reading, which is consistent with typical café behavior.}

\end{itemize}

\vspace{1ex}
\textbf{Weighted Average Score}: 0.8
}
\end{tcolorbox}
\end{tcolorbox}
% =========================
% Figure: PCR only
% =========================
\begin{tcolorbox}[enhanced,breakable,myboxstyle,title={High-Level Commonsense Reasoning}]
\begin{tcolorbox}[enhanced,breakable,drfstyle,title={PCR Evaluation}]
{\small

% =====================================================
% Causal Anticipation
% =====================================================
\noindent\textbf{\textcolor{purple}{Causal Anticipation}}
\begin{itemize}[leftmargin=15pt,itemsep=4pt]

  \item \textcolor{blue}{Q1}:  
  Does the person place the lid on the boiling pot correctly to cover it?  
  \\[-0.2em]
  \textbf{Ans:} Yes \quad
  {\color{gray}\textbf{Reason}: The person is seen holding the lid and placing it directly on top of the pot, covering the opening completely.}

  \item \textcolor{blue}{Q2}:  
  Does the person cover the pot before overflow?  
  \\[-0.2em]
  \textbf{Ans:} Yes \quad
  {\color{gray}\textbf{Reason}: The pot is already boiling with steam rising, and the person places the lid on it before any visible overflow occurs, indicating a preventive action.}

\end{itemize}

% =====================================================
% Preventive Timing
% =====================================================
\noindent\textbf{\textcolor{purple}{Preventive Timing}}
\begin{itemize}[leftmargin=15pt,itemsep=4pt]

  \item \textcolor{blue}{Q3}:  
  Does the lid appear securely placed on the pot, preventing steam from escaping visibly around the edges?  
  \\[-0.2em]
  \textbf{Ans:} Yes \quad
  {\color{gray}\textbf{Reason}: The lid is placed flat on the pot, and there is no visible steam escaping from the sides, suggesting a secure fit.}

  \item \textcolor{blue}{Q4}:  
  Are steam and splash effects physically coherent?  
  \\[-0.2em]
  \textbf{Ans:} Yes \quad
  {\color{gray}\textbf{Reason}: Steam is rising from the pot before the lid is placed, and the action of covering it is consistent with the goal of preventing splashing, which is a common physical effect of boiling liquids.}

  \item \textcolor{blue}{Q5}:  
  Does the lid stay in place and not slide off as the pot boils?  
  \\[-0.2em]
  \textbf{Ans:} Yes \quad
  {\color{gray}\textbf{Reason}: Once the lid is placed, it remains in position on the pot throughout the sequence, showing no signs of sliding or falling off.}

\end{itemize}

% =====================================================
% Visual Plausibility
% =====================================================
\noindent\textbf{\textcolor{purple}{Visual Plausibility}}
\begin{itemize}[leftmargin=15pt,itemsep=4pt]

  \item \textcolor{blue}{Q6}:  
  Are interactions between hand, lid, and pot plausible?  
  \\[-0.2em]
  \textbf{Ans:} Yes \quad
  {\color{gray}\textbf{Reason}: The hand movements are smooth and natural, with the person using both hands to carefully place the lid, which is a plausible interaction.}

  \item \textcolor{blue}{Q7}:  
  Is causal logic (boil → cover → calm) clearly visualized?  
  \\[-0.2em]
  \textbf{Ans:} Yes \quad
  {\color{gray}\textbf{Reason}: The video shows a clear causal sequence: the pot is boiling (cause), and the person covers it with a lid (preventive action), which logically leads to a more stable cooking process.}

  \item \textcolor{blue}{Q8}:  
  Is the prevention result (stable pot) shown?  
  \\[-0.2em]
  \textbf{Ans:} Yes \quad
  {\color{gray}\textbf{Reason}: The pot remains stable with the lid securely in place, and the steam is contained, demonstrating the successful prevention of splashing.}

\end{itemize}

\vspace{1ex}
\textbf{Weighted Average Score}: 1.00
}
\end{tcolorbox}
\end{tcolorbox}

\begin{tcolorbox}[enhanced,breakable,myboxstyle,title={High-Level Commonsense Reasoning}]
\begin{tcolorbox}[enhanced,breakable,drfstyle,title={CAR Evaluation}]
{\small
% =====================================================
% Action Sequence Coherence
% =====================================================

\noindent\textbf{\textcolor{purple}{Action Sequence Coherence}}
\begin{itemize}[leftmargin=15pt,itemsep=4pt]

  \item \textcolor{blue}{Q1}:  
  Is one worker demonstrating or pointing to parts of a machine?  
  \\[-0.2em]
  \textbf{Ans:} Yes \quad
  {\color{gray}\textbf{Reason}: The male worker is actively using his hands to point at and manipulate a component on the machine, demonstrating its parts and function to the female worker.}

  \item \textcolor{blue}{Q2}:  
  Does the worker actively use hand gestures to explain the machine's function?  
  \\[-0.2em]
  \textbf{Ans:} Yes \quad
  {\color{gray}\textbf{Reason}: The male worker uses his hands to gesture towards the machine's components and the area of operation, indicating a clear effort to explain the process to the female worker.}

\end{itemize}

% =====================================================
% Intentional Abstraction
% =====================================================

\noindent\textbf{\textcolor{purple}{Intentional Abstraction}}
\begin{itemize}[leftmargin=15pt,itemsep=4pt]

  \item \textcolor{blue}{Q3}:  
  Does the worker demonstrate proper safety procedures while explaining the machine's operation?  
  \\[-0.2em]
  \textbf{Ans:} Yes \quad
  {\color{gray}\textbf{Reason}: Both workers are wearing safety glasses and gloves, and the male worker is demonstrating the machine's operation in a controlled manner, which implies adherence to safety procedures.}

  \item \textcolor{blue}{Q4}:  
  Does the learner attempt to replicate or operate the machine afterward?  
  \\[-0.2em]
  \textbf{Ans:} Yes \quad
  {\color{gray}\textbf{Reason}: The female worker is seen placing her hands on the machine and mimicking the male worker's actions, indicating she is attempting to replicate the operation.}

  \item \textcolor{blue}{Q5}:  
  Do both exhibit engagement and focus on the same object?  
  \\[-0.2em]
  \textbf{Ans:} Yes \quad
  {\color{gray}\textbf{Reason}: Both individuals are focused on the same area of the machine, with their gazes and body postures directed towards the task, showing clear engagement.}

\end{itemize}

% =====================================================
% Contextual Emotional Coherence
% =====================================================

\noindent\textbf{\textcolor{purple}{Contextual Emotional Coherence}}
\begin{itemize}[leftmargin=15pt,itemsep=4pt]

  \item \textcolor{blue}{Q6}:  
  Do gestures or expressions indicate active teaching and learning?  
  \\[-0.2em]
  \textbf{Ans:} Yes \quad
  {\color{gray}\textbf{Reason}: The male worker's focused expression and demonstrative gestures, combined with the female worker's attentive posture and attempts to replicate, indicate an active teaching and learning dynamic.}

  \item \textcolor{blue}{Q7}:  
  Are machine controls and environment stable during the scene?  
  \\[-0.2em]
  \textbf{Ans:} Yes \quad
  {\color{gray}\textbf{Reason}: The machine and the surrounding environment remain static throughout the sequence, with no changes in the background or equipment, providing a stable context for the interaction.}

  \item \textcolor{blue}{Q8}:  
  Is there a concluding acknowledgment (e.g., nod, smile)?  
  \\[-0.2em]
  \textbf{Ans:} Yes \quad
  {\color{gray}\textbf{Reason}: In the final frames, the female worker is seen nodding her head and smiling, which serves as a clear acknowledgment of understanding or completion of the instruction.}

\end{itemize}
\vspace{1ex}
\textbf{Weighted Average Score}: 1.00
}
\end{tcolorbox}
\end{tcolorbox}
% =========================
% Figure: BBR only
% =========================
\begin{tcolorbox}[enhanced,breakable,myboxstyle,title={Embodied Physical Reasoning}]
\begin{tcolorbox}[enhanced,breakable,drfstyle,title={BBR Evaluation}]
{\small
% =====================================================
% Biomechanical Realism
% =====================================================
\noindent\textbf{\textcolor{purple}{Biomechanical Realism}}
\begin{itemize}[leftmargin=15pt,itemsep=4pt]

  \item \textcolor{blue}{Q1}:  
  Does the penguin push off with visible leg movement or body tilt to start sliding?  
  \\[-0.2em]
  \textbf{Ans:} Yes \quad 
  {\color{gray}\textbf{Reason}: The penguin is initially seen in a low, sliding posture. In the later frames, it begins to lift its body and extend its legs, indicating a push-off motion to initiate movement, which is a common behavior for penguins transitioning from sliding to walking.}

  \item \textcolor{blue}{Q2}:  
  Is sliding speed gradually decreasing instead of abrupt stops?  
  \\[-0.2em]
  \textbf{Ans:} Yes \quad 
  {\color{gray}\textbf{Reason}: The penguin's motion shows a clear deceleration. It starts with a more dynamic, forward-leaning posture and gradually slows down, with its legs becoming more engaged in walking, indicating a gradual decrease in speed rather than an abrupt stop.}

\end{itemize}

% =====================================================
% Environmental Interaction
% =====================================================
\noindent\textbf{\textcolor{purple}{Environmental Interaction}}
\begin{itemize}[leftmargin=15pt,itemsep=4pt]

  \item \textcolor{blue}{Q3}:  
  Does belly and flipper contact remain consistent with the surface?  
  \\[-0.2em]
  \textbf{Ans:} Yes \quad 
  {\color{gray}\textbf{Reason}: The penguin maintains consistent contact with the ice surface throughout the sequence. Its belly remains in contact with the ice during the initial slide, and its feet are in contact during the walking phase, which is consistent with the described behavior.}

  \item \textcolor{blue}{Q4}:  
  Does the penguin’s body maintain a low, horizontal posture throughout the slide?  
  \\[-0.2em]
  \textbf{Ans:} Yes \quad 
  {\color{gray}\textbf{Reason}: The penguin maintains a low, horizontal posture during the slide, which is characteristic of penguin locomotion on ice. This posture is evident in the initial frames and is maintained as it transitions to walking.}

  \item \textcolor{blue}{Q5}:  
  Do flippers adjust slightly to maintain balance during motion?  
  \\[-0.2em]
  \textbf{Ans:} Yes \quad
  {\color{gray}\textbf{Reason}: The penguin's flippers are visibly adjusting their position throughout the sequence. They are held out to the sides for balance during the slide and are seen moving as the penguin transitions to walking, indicating an active role in maintaining stability.}

\end{itemize}

% =====================================================
% Ecological Coherence
% =====================================================
\noindent\textbf{\textcolor{purple}{Ecological Coherence}}
\begin{itemize}[leftmargin=15pt,itemsep=4pt]

  \item \textcolor{blue}{Q6}:  
  Is snow or ice displaced subtly beneath or behind the body?  
  \\[-0.2em]
  \textbf{Ans:} No \quad 
  {\color{gray}\textbf{Reason}: There is no visible displacement of snow or ice beneath or behind the penguin. The surface appears smooth and undisturbed, which is not typical for a sliding motion that would create a trail or disturb the surface.}

  \item \textcolor{blue}{Q7}:  
  Does the penguin's body tilt backward or forward as it slides to a stop?  
  \\[-0.2em]
  \textbf{Ans:} Yes \quad 
  {\color{gray}\textbf{Reason}: As the penguin slows down, its body tilts forward, particularly in the later frames. This forward tilt is consistent with the momentum of the slide and the effort to maintain balance as it comes to a stop.}

  \item \textcolor{blue}{Q8}:  
  Does the penguin's body maintain a consistent posture without wobbling excessively as it slows down?  
  \\[-0.2em]
  \textbf{Ans:} Yes \quad 
  {\color{gray}\textbf{Reason}: The penguin's body maintains a relatively stable and consistent posture throughout the motion. While there are adjustments for balance, there is no excessive wobbling, indicating a controlled and stable transition from sliding to walking.}

\end{itemize}

\vspace{1ex}
\textbf{Weighted Average Score}: 0.9
}
\end{tcolorbox}
\end{tcolorbox}

% =========================
% Figure: CAC only
% =========================
\begin{tcolorbox}[enhanced,breakable,myboxstyle,title={Embodied Physical Reasoning}]
\begin{tcolorbox}[enhanced,breakable,drfstyle,title={CAC Evaluation}]
{\small
% =====================================================
% Temporal Synchronization
% =====================================================
\noindent\textbf{\textcolor{purple}{Temporal Synchronization}}
\begin{itemize}[leftmargin=15pt,itemsep=4pt]

  \item \textcolor{blue}{Q1}:  
  Is the gamer actively interacting with game controls?  
  \\[-0.2em]
  \textbf{Ans:} Yes \quad
  {\color{gray}\textbf{Reason}: The gamer's hands are positioned on a controller, and his fingers are actively pressing buttons, indicating direct interaction with the game controls.}

  \item \textcolor{blue}{Q2}:  
  Is speech activity occurring through visible mouth motion?  
  \\[-0.2em]
  \textbf{Ans:} Yes \quad
  {\color{gray}\textbf{Reason}: The gamer's mouth is visibly open and moving in several frames, consistent with speech activity, which is necessary for communicating with teammates.}

\end{itemize}

% =====================================================
% Attention Coordination
% =====================================================
\noindent\textbf{\textcolor{purple}{Attention Coordination}}
\begin{itemize}[leftmargin=15pt,itemsep=4pt]

  \item \textcolor{blue}{Q3}:  
  Are speech and game actions overlapping naturally?  
  \\[-0.2em]
  \textbf{Ans:} Yes \quad
  {\color{gray}\textbf{Reason}: The actions of holding the controller and speaking occur simultaneously throughout the sequence, which is a natural and expected overlap in a gaming environment.}

  \item \textcolor{blue}{Q4}:  
  Does gaze remain fixed on the game screen?  
  \\[-0.2em]
  \textbf{Ans:} Yes \quad
  {\color{gray}\textbf{Reason}: The gamer's head is turned towards the monitor, and his gaze is directed forward, consistent with focusing on the game screen during play.}

  \item \textcolor{blue}{Q5}:  
  Is hand–eye coordination consistent with gameplay timing?  
  \\[-0.2em]
  \textbf{Ans:} Yes \quad
  {\color{gray}\textbf{Reason}: The continuous and fluid motion of the gamer's hands on the controller, combined with his focused expression, suggests consistent hand-eye coordination appropriate for real-time gameplay.}

\end{itemize}

% =====================================================
% Physical Realism
% =====================================================
\noindent\textbf{\textcolor{purple}{Physical Realism}}
\begin{itemize}[leftmargin=15pt,itemsep=4pt]

  \item \textcolor{blue}{Q6}:  
  Does the video show the gamer’s mouth movements synchronized with the spoken dialogue?  
  \\[-0.2em]
  \textbf{Ans:} Yes \quad
  {\color{gray}\textbf{Reason}: The movement of the gamer's mouth, including opening and closing, is clearly visible and synchronized with the implied speech, indicating that his mouth movements are consistent with spoken dialogue.}

  \item \textcolor{blue}{Q7}:  
  Is voice–action synchronization plausible for real gaming?  
  \\[-0.2em]
  \textbf{Ans:} Yes \quad
  {\color{gray}\textbf{Reason}: The simultaneous actions of playing a game and communicating via a headset are a standard and plausible scenario in competitive gaming, where real-time coordination is essential.}

  \item \textcolor{blue}{Q8}:  
  Is overall tempo natural without temporal conflict?  
  \\[-0.2em]
  \textbf{Ans:} Yes \quad
  {\color{gray}\textbf{Reason}: The overall tempo of the actions, including the natural flow of speech and controller inputs, is consistent and free of any temporal conflicts or unnatural pauses.}

\end{itemize}

\vspace{1ex}
\textbf{Weighted Average Score}: 1.00
}
\end{tcolorbox}
\end{tcolorbox}

\end{document}